\documentclass[11pt,a4paper]{article}
\usepackage{times,latexsym}
\usepackage{url}
\usepackage[T1]{fontenc}

\usepackage[acceptedWithA]{tacl2018v2}

\usepackage{CJKutf8}

\usepackage{times}
\usepackage{latexsym}
\usepackage{booktabs}
\usepackage{multirow}
\usepackage{multicol}
\usepackage[utf8]{inputenc}
\usepackage{pifont}
\usepackage{soul}
\usepackage{color}
\usepackage{comment}
\newcommand{\hlred}{\colorlet{c}{red!20}\sethlcolor{c}\hl}
\newcommand{\hlgreen}{\colorlet{c}{green!20}\sethlcolor{c}\hl}

\usepackage{bm}

\usepackage{graphics}
\usepackage{graphicx}
\usepackage{subfigure}
\usepackage{CJK}
\usepackage{amsmath}
\usepackage{amssymb}
\usepackage{xspace,mfirstuc,tabulary}

\newif\iftaclinstructions
\taclinstructionsfalse 
\iftaclinstructions
\renewcommand{\confidential}{}
\renewcommand{\anonsubtext}{(No author info supplied here, for consistency with
TACL-submission anonymization requirements)}
\newcommand{\instr}
\fi

\iftaclpubformat 

\else

\fi

\title{LOT: A Story-Centric Benchmark for Evaluating Chinese Long Text Understanding and  Generation}

\author{
 Template Author\Thanks{The {\em actual} contributors to this instruction
 document and corresponding template file are given in Section
 \ref{sec:contributors}.} \\
 Template Affiliation/Address Line 1 \\
 Template Affiliation/Address Line 2 \\
 Template Affiliation/Address Line 2 \\
  {\sf template.email@sampledomain.com} \\
}

\author{
 Jian Guan$^1$, Zhuoer Feng$^1$, Yamei Chen$^1$, Ruilin He$^2$,\\\textbf{Xiaoxi Mao$^3$, Changjie Fan$^3$ and Minlie Huang}$^1$\Thanks{~Corresponding author}\\
\small{$^1$The CoAI group, DCST; $^1$Institute for Artificial Intelligence; $^1$State Key Lab of Intelligent Technology and Systems;}\\
\small{$^1$Beijing National Research Center for Information Science and Technology;} 
\small{$^1$Tsinghua University, Beijing 100084, China.}\\
\small{$^2$Huawei Technologies Co., Ltd.  $^3$Netease Fuxi AI Lab.}\\
  \small{\texttt{\{j-guan19,fze17\}@mails.tsinghua.edu.cn}, \texttt{chenziym4132013@163.com, heruilin@huawei.com}}, \\\small{ \texttt{ \{maoxiaoxi,fanchangjie\}@corp.netease.com}}, \small{\texttt{aihuang@tsinghua.edu.cn}} \\
  
}
\date{}

\begin{document}
\maketitle
\begin{abstract}
Standard multi-task benchmarks are essential for developing pretraining models that can generalize to various downstream tasks. Existing benchmarks for natural language processing~(NLP) usually focus only on understanding or generating short texts. However, long text modeling requires many distinct abilities in contrast to short texts, such as the modeling of long-range discourse and commonsense relations, and the coherence and controllability of generation. The lack of standardized benchmarks makes it difficult to assess these abilities of a model and fairly compare different models, especially Chinese models. Therefore, we propose a story-centric benchmark named LOT for evaluating Chinese long text modeling, which aggregates two understanding tasks and two generation tasks. We construct new datasets for these tasks based on human-written Chinese stories with hundreds of words. Furthermore, we release an encoder-decoder-based Chinese long text pretraining model named LongLM with up to 1 billion parameters. 
We pretrain LongLM on 120G Chinese novels with two generative tasks  including text infilling and conditional continuation. Extensive experiments show that LongLM outperforms similar-sized pretraining models substantially on both the understanding and generation tasks in LOT. 

\end{abstract}

\section{Introduction}

\begin{table}[!t]
\footnotesize
    \centering
    \begin{tabular}{p{207pt}}
    \toprule
    Effendi’s son is \textbf{eccentric}, always \textbf{behaving opposed to what Effendi has ordered him to do}. Familiar to his son's temper, Effendi usually \textbf{communicates using irony}. One day, the father and son were blocked by a river after purchasing flour from a mill. And while they were crossing the river, one bag on the donkey's back lost its weight and leaned. Effendi told his son with \textbf{irony}:``My boy! \textbf{drop the sack into the river}!'' The son heard the words and thought:``I have been opposed to my father for so many years. For this only time, I have to \textbf{obey} him.'' Therefore, he followed Effendi's words and indeed \textbf{pushed the sack into the river}. ``My boy! What are you doing?”Effendi shouted in \textbf{anger}.'' ...\\
   \bottomrule
    \end{tabular}
    \caption{A long text example. 
  The concepts and events concerning commonsense and discourse relations are highlighted in \textbf{bold}.} 
    \label{tab:example}
\end{table}

Pretrained language models have achieved significant advances in various natural language understanding~(NLU) and generation~(NLG) tasks~\cite{devlin2018bert,radford2019language}. Standard benchmarks such as GLUE~\cite{wang2018glue} further boost the improvement and fast iteration of pretrained models. Popular benchmarks usually aggregate multiple tasks to spur the progress of generalizable models. But these benchmarks focus mainly on understanding or generating short texts. For example, the GLUE tasks take at most two sentences as input. 
And most tasks in NLG benchmarks such as GLGE~\cite{liu2020glge} and GEM~\cite{gehrmann2021gem} 
require generating only several words~(e.g., dialogue generation). 
Although there have been many models pretrained on long texts such as
GPT3~\cite{brown2020language} and CPM~\cite{zhang2020cpm}), 
the lack of benchmark datasets makes it difficult to fully assess and compare their abilities of long text modeling. 

In this paper, we present LOT, a benchmark for evaluating \textit{Chinese \textbf{LO}ng \textbf{T}ext understanding and generation}. As shown in Table~\ref{tab:example}, modeling long texts requires many distinct abilities compared to short texts,
including (1) commonsense reasoning regarding characters' reaction and intention, 
and knowledge about physical objects~(e.g., ``river'') and abstract concepts~(e.g., ``irony''); 
(2) modeling discourse-level features such as inter-sentence relations~(e.g., causality) and global discourse structures~(e.g., the order of events);
and (3) the generation coherence and controllability, which require both maintaining a coherent plot and adhering to controllable attributes~(e.g., topics). 
Accordingly, LOT contains two understanding tasks and two generation tasks regarding the above 
abilities. 
We construct new datasets for these tasks based on various kinds of stories such as fables and fairy tales collected from public web resources, considering that stories usually contain abundant commonsense and discourse relations. All these tasks require processing stories with hundreds of words. Note that LOT does not involve extra-long texts with thousands of words since the complicated linguistic phenomena in these texts make it hard to test individual abilities and guide the improvement of generation models.

Furthermore, we release LongLM, a \textit{Chinese \textbf{Long} text pretraining \textbf{L}anguage \textbf{M}odel}. 
LongLM is a Transformer-based model with an encoder-decoder architecture. LongLM has three different versions ranging from 60 million to 1 billion parameters. We pretrain LongLM on 120G Chinese novels 
with two generative tasks, including text infilling~\cite{bart} and conditional continuation~\cite{radford2018improving}. The pretraining data do not include other types of texts~(e.g., news, Wiki-texts) since we mainly focus on commonsense and discourse relations within general long texts instead of factual and technical knowledge. To the best of our knowledge, LongLM is the first pretraining model of the same size scale that focuses on modeling long-form stories. 
Extensive experiments on LOT show that LongLM outperforms strong baselines substantially on both the understanding and generation tasks. However, we also observe that LongLM is still far behind human performance, which requires better semantic representations of events and deeper modeling of the commonsense and discourse relations between them. 
We summarize the main contributions of this paper as follows:

\noindent\textbf{I.} We propose a new story-centric benchmark LOT for evaluating Chinese long text understanding and generation. LOT consists of four tasks for testing the fundamental abilities to model long texts. We also present new datasets for these tasks.

\noindent\textbf{II.} We release a new Chinese pretraining model named LongLM. Experiment results demonstrate the strong performance of LongLM on LOT, but there still exists huge room for improvement 
\footnote{The LOT benchmark, the pretraining resources and the appendix are available at \url{https://github.com/thu-coai/LOT-LongLM}.}

\section{Related Work}

\paragraph{NLP Benchmarks} Recently, there have been a lot of multi-task benchmarks proposed to drive the progress of generalizable models. The benchmarks usually aggregate multiple model-agnostic tasks under a unified framework, enabling researchers to fairly compare different models. SentEval~\cite{conneau2018senteval} gathered multiple classification tasks involving either one or two sentences as inputs to evaluate sentence representations. DiscoEval~\cite{chen2019evaluation} extended these tasks to the discourse level regarding inter-sentence relations. GLUE~\cite{wang2018glue} included more diverse tasks such as natural language inference~\cite{DBLP:journals/corr/RocktaschelGHKB15}. 
\citet{sarlin2020superglue} proposed SuperGLUE as a more challenging counterpart of GLUE by introducing multi-sentence tasks. But the additional tasks are only limited to the formats of coreference resolution and question answering. 
In addition to these English benchmarks, many benchmarks were proposed to evaluate NLU for other languages such as CLUE~\cite{xu2020clue} for Chinese. 
Moreover, GLGE~\cite{liu2020glge} and GEM~\cite{gehrmann2021gem} were proposed for evaluating NLG models across diversified generation tasks such as text summarization and personalizing dialogue. However, 
there is no benchmark designed specifically for long text modeling, especially Chinese. Additionally, the above benchmarks were originally designed to cover as diverse task formats as possible. In contrast, we design the LOT tasks with the guidance of necessary abilities for long text modeling as suggested by \citet{ribeiro-etal-2020-beyond}, making it easier to figure out where models are failing, and how to improve them. 
 
\begin{table*}[!t]
\scriptsize
    \centering
    \begin{tabular}{l|p{88pt}|p{138pt}|p{95pt}|p{42pt}}
    \toprule
\textbf{Tasks}&\textbf{Abilities}&\textbf{Inputs}&\textbf{Outputs}&\textbf{Metrics}\\
\midrule
\textbf{ClozeT}&Commonsense Reasoning&A text with a sentence removed~(the position specified); Two candidate sentences.&Choosing the correct sentence from two candidates.&Accuracy\\
\midrule
\textbf{SenPos}&Inter-sentence Relationship&A text with a sentence removed~(the position unspecified); The removed sentence.&Choosing the correct position for the removed sentence.&Accuracy\\
\midrule
\midrule
\textbf{PlotCom}&Commonsense Reasoning; Inter-sentence Relationship&A text with a sentence removed~(the position specified).&Generating a sentence to complete the text.&BLEU; Dist \\
\midrule
\textbf{OutGen}&Discourse Structure;\newline Coherence; Controllability &{A title, an outline as an out-of-order set of phrases about characters and events.}&{Generating a coherent text adhering to the title and outline.}&BLEU; Dist;\newline Cover; Order  \\
   \bottomrule
    \end{tabular}
    \caption{Overview of the tasks in LOT for the abilities they test, inputs and outputs, and the evaluation metrics. \textbf{Dist} and \textbf{Cover} refer to Distinct and Coverage~(Section \ref{sec:metrics}), respectively.}
    \label{tab:desc}
\end{table*}

\paragraph{Long Text Datasets} Previous studies in the field of long text modeling have frequently focused on the ROCStories~\cite{mostafazadeh2016corpus} and WritingPrompts~\cite{fan2018hierarchical} datasets. ROCStories contains 100k artificial five-sentence stories, 
while WritingPrompts consists of 300K pairs of prompts and stories with hundreds of words.
Recent works collected stories with thousands of words 
to model longer-range dependencies, such as WikiText-103~\cite{merity2016pointer}, 
{roleplayerguild}~\cite{louis2018deep}, PG-19~\cite{Rae2020Compressive}, {STORIUM}~\cite{akoury2020storium} and Long-Range Arena~\cite{tay2020long}. However, these datasets are written in English. 
LOT will drive the development of Chinese language models. 

Moreover, LOT does not include datasets of extra-long texts like PG-19 for the following two reasons: (1) Extra-long texts are far beyond the scope of current machine learning models because the discourse-level linguistic phenomena are entangled and complicated in these texts. Therefore, extra-long texts usually serve for computing perplexity of language models~\cite{dai2019transformer} but hardly provide fine-grained guidance for improving model designs. (2) LOT aims not to spur research on building fuller connections across tokens within an extra-long sequence, but to drive the progress of machines in the aforementioned fundamental abilities for long text modeling. 


\paragraph{Story Understanding and Generation} 

LOT is centered on fundamental abilities for long text modeling and thus includes four story understanding and generation tasks concerning commonsense and discourse relations. 
Recent studies have proposed various tasks to evaluate story understanding and generation. 
Firstly, story ending selection~\cite{mostafazadeh2016corpus}, story ending generation~\cite{guan2019story} and story completion~\cite{DBLP:conf/ijcai/Wang019b} focused on the {commonsense reasoning} ability on inter-event causal and temporal relations. Secondly, \citet{chen2019evaluation} evaluated the ability to model discourse relations by predicting the position of a sentence or a paragraph in a text. Thirdly, some works focused on the {coherence} of story generation conditioned on short prompts~\cite{fan2018hierarchical}, titles~\cite{yao2019plan} and beginnings~\cite{guan2020knowledge}. 
Fourthly, some studies centered on {controllability}, i.e., the imposing of controllable attributes on story generation such as keywords~\cite{DBLP:conf/emnlp/XuPSPFAC20}, emotional trajectories~\cite{brahman2020modeling}, outlines~\cite{rashkin2020plotmachines} and styles~\cite{kong-etal-2021-stylized}. LOT is a comprehensive benchmark to test the above abilities for Chinese long text modeling.

On the other hand, LOT does not involve those 
tasks that require learning more particular features of stories, such as event chains~\cite{chambers2008unsupervised}, character types~\cite{bamman2013learning}, inter-character relations~\cite{chaturvedi2016modeling,chaturvedi2017unsupervised}, social networks~\cite{agarwal2013automatic} and abstractive structures~\cite{finlayson2012learning}. 
Non-neural story generation models usually retrieved events from a knowledge base with pre-specified semantic relations based on handcrafted rules~\cite{li2013story}, which are costly and lack generalization. In this paper, we focus mainly on evaluating neural models for story understanding and generation.

\section{LOT Benchmark}
\begin{table}[!t]
\scriptsize
    \centering
    \begin{tabular}{l|ccc}
    \toprule
\textbf{Datasets}&\textbf{Train}&\textbf{Val}&\textbf{Test}\\
\midrule
\midrule
\multicolumn{4}{c}{\textbf{Task:} \textbf{ClozeT}}\\
\midrule
\textbf{\# Examples}\ & 644 & 294 & 294\\
\textbf{Vocabulary Size}\ &  9k& 7k & 7k\\
\midrule
\textbf{Avg. \# Char in Input Text}&139.07&138.95&141.15\\
\textbf{Avg. \# Word in Input Text}\ &89.28&89.03&90.20 \\
\textbf{Avg. \# Sent in Input Text}\ &5.95&5.94&5.95 \\
\midrule
\textbf{Avg. \# Word in Candidate}\ & 15.60&16.38&15.75\\
\midrule
\midrule
\multicolumn{4}{c}{\textbf{Task:} \textbf{SenPos}}\\
\midrule
\textbf{\# Examples} & 20,000 & 800 & 863\\
\textbf{Vocabulary Size} & 147k & 10k & 22k\\
\midrule
\textbf{Avg. \# Char in Input Text}&289.59&258.48&258.52\\
\textbf{Avg. \# Word in Input Text} & 254.11 & 224.20 & 223.25\\
\textbf{Avg. \# Sent in Input Text} & 9.61 & 8.43 & 8.44\\
\textbf{Avg. \# Word in Removed
Sent} & 30.48 & 29.28 & 30.26\\
\midrule
\textbf{Avg. \# Candidate Positions} & 8.05 & 6.91 & 6.91\\
\midrule
\midrule
\multicolumn{4}{c}{\textbf{Task:} \textbf{PlotCom}}\\
\midrule
\textbf{\# Examples} & 13,099 & 465 & 464\\
\textbf{Vocabulary Size} & 22k & 8k & 8k\\
\midrule
\textbf{Avg. \# Char in Input Text}&164.35&137.67&133.26\\
\textbf{Avg. \# Word in Input Text} & 105.48 & 87.56 & 84.98\\
\textbf{Avg. \# Sent in Input Text} & 7.17 & 5.59 & 5.48\\
\midrule
\textbf{Avg. \# Word in Output Sent} & 15.08 & 15.96 & 16.15\\
\midrule
\midrule
\multicolumn{4}{c}{\textbf{Task:} \textbf{OutGen}}\\
\midrule
\textbf{\# Examples}& 1,456 & 242 & 729\\
\textbf{Vocabulary Size} & 19k  & 6k & 12k\\
\midrule
\textbf{Avg. \# Word in Input Title} & 4.64&4.89&4.64\\
\textbf{Avg. \# Word in Input Outline} & 19.20  & 19.05 & 19.47\\
\textbf{Avg. \# Phrase in Input Outline} & 8.00  & 8.00 & 8.00\\
\midrule
\textbf{Avg. \# Char in Output Text}&169.94&169.80&170.49\\
\textbf{Avg. \# Word in Output Text} & 108.91  & 108.68 & 109.04\\
\textbf{Avg. \# Sent in Output Text} & 7.20  & 7.11 & 7.15\\
\bottomrule
    \end{tabular}
    \caption{Data statistics of LOT tasks. The abbreviation \textbf{char}/\textbf{sent}/\textbf{len} is short for \textbf{character}/\textbf{sentence}/\textbf{length}, respectively.}
    \label{tab:stat}
\end{table}
We design LOT as an aggregation of two understanding tasks including \textit{Cloze Test}~(ClozeT) and \textit{Sentence Position Prediction}~(SenPos), and two generation tasks including \textit{Plot Completion}~(PlotCom) and \textit{Outline-conditioned Generation}~(OutGen). We show the task descriptions and data statistics in Table~\ref{tab:desc} and \ref{tab:stat}, respectively. We use the \texttt{jieba} tokenizer\footnote{\url{https://github.com/fxsjy/jieba}} for word tokenization. 

We design  LOT 
based on the following principles: 
\textbf{(1) Task Diversity:} The tasks vary in task formats, types and lengths of inputs and outputs, focused abilities, making LOT a comprehensive framework for evaluating the generalization of models.
\textbf{(2) Task Difficulty:} The tasks take hundreds of words as inputs or outputs, and do not involve domain-specific knowledge about science, films, etc. Therefore, they are beyond the scope of current state-of-the-art models, but are solvable by most Chinese native speakers.
\textbf{(3) Task Formulation:} The tasks have been well formulated in prior studies and agreed to be challenging but meaningful. We introduce new Chinese datasets for these tasks, which are constructed to focus more specifically on testing a certain ability than original datasets.
\textbf{(4) Automatic Evaluation:} These tasks have reliable automatic metrics to evaluate the focused abilities.
We exclude open-ended generation tasks such as story generation from titles, which is difficult to automatically evaluate~\cite{guan2021openmeva} since the tasks suffer from the notorious one-to-many issue: there are many plausible outputs for the same input~\cite{zhao2017learning}. 

We constructed datasets for LOT through automatic and manual annotation. Firstly, we crawled human-written stories from public web pages as the data source. 
These stories are under licenses that allow use and redistribution for research purposes. 
Then, we hired a commercial team to create the LOT examples. The team is led by a professional screenwriter and has taken on hundreds of NLP annotation projects. All annotators are native Chinese speakers and well-trained for the annotation tasks. We show the full list of the source web pages and the annotation details in the appendix.



\subsection{Cloze Test}
\citet{mostafazadeh2016corpus} introduced the Story Cloze Test~(SCT) task 
for evaluating story comprehension, 
which requires selecting the right ending from two candidates for a four-sentence leading context. 
However, SCT suffers from the following issues: (1) Its dataset is artificial 
and contains innate biases between right and wrong endings in some features such as lengths~\cite{schwartz2017effect,sharma2018tackling}. Such biases may leak information about the target labels. (2) SCT focuses on reasoning only endings but neglects other types of reasoning, such as abductive reasoning~\cite{bhagavatula2019abductive}, which requires reasoning what happens between observed beginnings and endings. (3) SCT limits the scope of commonsense reasoning to realistic events. 
The limitation may be neither necessary nor sufficient. For example, {``Cupid can fly''} can be reasoned based on common sense although it is not realistic, while some story settings may be realistic but fail to be reasoned only based on the context and common sense, as shown in Table~\ref{tab:cs}. Therefore, when constructing our ClozeT dataset, we adopt the following approaches to alleviate the above issues: (1) All examples are derived from existing human-written stories.  (2) We allow annotators to create examples where the removed sentence is initially in the middle of the story. (3) We change the scope of commonsense reasoning to all events that embody characters' reaction and intention, or the nature of physical objects and concepts. 
Table~\ref{tab:clozetest} shows two ClozeT examples. Furthermore, we also conducted experiments to investigate the potential biases of our dataset in Section \ref{bias_invest}.

\begin{table}[!t]
\scriptsize
    \centering
    \begin{tabular}{p{205pt}}
    \toprule
A goblin had buried a treasure under the ground. \textit{\hlgreen{After that, he received a long flight mission from the Devil King.}} The goblin began to worry about how to guard the treasure during his mission. \textbf{\hlred{The goblin thought for a long time and decided to give the treasure to a miser.}} {The miser clung to his vault even when he was asleep, so the goblin trusted him very much} $\cdots$\\
\bottomrule
    \end{tabular}
    \caption{An example for selecting a sentence that can be reasoned based on the context and common sense~(in \hlred{\textbf{red}}). We also highlight a sentence that does not satisfy the requirement in \hlgreen{\textit{green}}, which introduces a new character \textit{``the Devil King''}.} 
    \label{tab:cs}
\end{table}



\paragraph{Story Filtering} 
To ensure the quality of LOT examples, we asked annotators to judge whether each crawled story meets the following definition: ``anything which is told in the form of a coherent event sequence involving several specific and related characters''~\cite{mostafazadeh2016corpus}. 
We provided detailed cases for annotators to instruct them about this definition. 
Then, annotators needed to refine those stories which do not meet the definition by rewriting the plots. 
They should also clean up the stories by the following heuristics:
(1) refusing examples which may violate ethical principles (e.g., discrimination); (2) deleting noisy words (e.g., links); (3) changing slang and informal words into standard modern Chinese; (4) rewriting all dialogues to objective events.
Finally, we collected 2,427 high-quality Chinese stories, which will be used to construct the datasets for the ClozeT, PlotCom and OutGen tasks. 

\paragraph{Dataset Construction} We presented the stories to another group of annotators to construct the ClozeT dataset. For each story, they should select a sentence as the right candidate that can be reasoned based on the context and common sense. Table~\ref{tab:cs} shows an example presented to the annotators to illustrate how to judge whether a sentence satisfies this requirement. 
Then, the annotators rewrite the sentence into another one as the wrong candidate that maintains a good topical relatedness with the context but violates common sense. The wrong candidates should either embody unreasonable reactions or intentions, or violate the nature of physical objects or concepts. 
And we require annotators not to select 
the first sentence, which usually aims to introduce story settings 
instead of narrating an event. We browse through the annotation results and give the annotators detailed feedback before approving their submissions. Finally, we collected 1,232 examples in total and split them for training, validation and testing.

\subsection{Sentence Position Prediction}

We use the sentence position prediction task~\cite{chen2019evaluation} to evaluate the ability to capture inter-sentence relations~(e.g., causality). We formulate the task as follows: given a text with a sentence removed, models should choose the correct position of the sentence in the text from multiple candidates. \citet{chen2019evaluation} constructed an English dataset for this task by randomly removing sentences from existing texts. However, such examples may be invalid since a sentence may have multiple plausible positions in a text, as illustrated in Table~\ref{tab:case_poor_ssp}. 
Therefore, we construct the dataset for our task based on the following pipeline: (1) extracting paragraphs with less than 500 words from crawled stories; (2) randomly selecting a sentence to remove for each paragraph, 
and regarding all positions between two adjacent sentences as candidates\footnote{We set the minimum length of the removed sentence to 10 Chinese characters, and we merge a sentence in a story with its neighbors if it contains less than 10 characters.}, and (3) asking annotators to refine part of the auto-constructed examples as the validation and test sets, and the remaining as the training set.  Table~\ref{tab:case_ssp} shows two SenPos examples.

\begin{table}[!ht]
\scriptsize
    \centering
    \begin{tabular}{p{205pt}}
    \toprule
\textbf{Text:} I couldn't control my anger very well.\texttt{[1]}My parents would yell at me, and i ran to my room.\texttt{[2]}I buried my head in a pillow and screamed.\texttt{[3]}I threw my pillow and hit it hard.\\
\midrule
\textbf{Removed Sentence:} {{{I tried to express my anger.}}}\\
\bottomrule
    \end{tabular}
    \caption{A poor example for the SenPos task. 
    The removed sentence has multiple reasonable positions including \texttt{[2]} and \texttt{[3]} in the original text.}
    \label{tab:case_poor_ssp}
\end{table}

\begin{table*}[!t]
\scriptsize
    \centering
    \begin{tabular}{p{276pt}p{73pt}p{70pt}}
    \toprule
    \textbf{Context}&\textbf{Wrong Candidates}&\textbf{Right Candidates}\\
    \midrule
    \begin{CJK}{UTF8}{gbsn}\tiny {{傻}}狼和狐狸偷来一罐蜂蜜藏在树洞里，它俩规定谁也不许偷吃。可狐狸第二天就把蜂蜜偷吃光了。傻狼几次找狐狸去吃蜂蜜，狐狸总是不去。傻狼实在忍不住，结果跑过去一看，蜂蜜竟然见了底。傻狼心里懊恼，它想着，都怪狐狸不来吃，蜂蜜全都干掉了，丝毫没有怀疑狐狸。\texttt{[MASK]}\end{CJK}&\begin{CJK}{UTF8}{gbsn}\tiny 狐狸听说后非常生气，他再也不跟傻狼一起找吃的了。\end{CJK}&\begin{CJK}{UTF8}{gbsn}\tiny 狐狸听说后，更加积极地跟傻狼一起去找吃的了。\end{CJK}\\
    A \hlred{\textbf{silly}} wolf and a fox stole a jar of honey and then hid it in a tree hole. They agreed that neither of them were allowed to eat the honey alone. However, \hlred{\textbf{the fox sneaked back to eat up all the honey the next day}}. Afterwards, whenever the wolf asked the fox to eat the honey together, the fox always refused its request. Finally the wolf could not help coming back to the tree hole and found that the jar had been empty. The wolf felt very regretful that the honey \hlred{\textbf{became dry because it had been too long}}. \hlred{\textbf{It had no doubts about the fox at all.}} \texttt{[MASK]}& When hearing this, \hlgreen{\textit{the fox became very angry}} and decided \hlgreen{\textit{no longer to look for food together with the wolf}}.&After hearing this, the fox became \hlred{\textbf{more active to look for food together with the wolf.}}\\
    \midrule
    \begin{CJK}{UTF8}{gbsn}\tiny 从前，山脚下住着母子二人。儿子长大后出门学艺，一直没有回来。妈妈就到城里找。哪知道儿子当了官竟不认她了。老妈妈坐在路边伤心地哭了，一个青年路过，知道了原由，将她接回家里。\texttt{[MASK]}他下令将那个不孝顺的儿子贬为了平民。而他的妈妈则在王宫里过上了幸福的生活。\end{CJK}&\begin{CJK}{UTF8}{gbsn}\tiny 谁知，这个青年也是一个官。 \end{CJK}&\begin{CJK}{UTF8}{gbsn}\tiny 谁知，这个青年竟是王子。\end{CJK}\\
    Once upon a time, there lived a mother and her son at the foot of a mountain. After her son grew up, he went out to learn skills and never came back. Therefore, the mother went to the nearby city to look for him. However, \hlred{\textbf{her son became an official}} and disowned his mother. The mother sat by the roadside and cried sadly. A young man passed by and knew the cause. Then the man \hlred{\textbf{took her home}}. \texttt{[MASK]} He \hlred{\textbf{decreed}} to remove the position of the disobedient son. And the mother lived happily in the \hlred{\textbf{palace}}.&Actually the man was also an \hlgreen{\textit{official}}. &Actually the man was the \hlred{\textbf{prince}} of the city.\\
   \bottomrule
    \end{tabular}
    \caption{Two ClozeT examples. The right candidates are extracted from the original stories~(at the position of ``\texttt{[MASK]}'') while the wrong candidates are written by crowd-sourced annotators. The first example focuses on common sense regarding the \textit{fox}'s reaction to the \textit{silly wolf}'s behaviour, while the second example focuses on common sense regarding the relations between \textit{palace} and \textit{prince}. We highlight the entities and events related to the commonsense relations in \hlred{\textbf{red}}, and those which violate common sense in the wrong candidates in \hlgreen{\textit{green}}.}
    \label{tab:clozetest}
\end{table*}

\begin{table*}[!ht]
\scriptsize
    \centering
    \begin{tabular}{p{329pt}p{67pt}p{23pt}}
    \toprule
    \textbf{Texts}&\textbf{Removed Sentences}&\textbf{Labels}\\
    \midrule
        \begin{CJK}{UTF8}{gbsn}\tiny \textbf{有一个}姓蒋的人，祖父和父亲在捕蛇的时候被蛇咬死了，但是他却继续捕蛇。\texttt{[1]} 当柳宗元劝他不要在捕蛇的时候，这个人竟大哭起来，宁愿被蛇咬死，也不愿意放弃捕蛇。\texttt{[2]} 有的乡亲早已倾家荡产，食不裹腹了。\texttt{[3]} 差役们到进村子里收税赋的时候，横冲直撞，粗声叫骂，大打出手，乡亲们胆战心惊，苦苦哀求。\texttt{[4]} 这种场面连鸡狗都得不到安宁，何况人呢！\end{CJK}&\begin{CJK}{UTF8}{gbsn}\tiny 因为他必须靠捕蛇才能上缴官府的赋税。\end{CJK}&\texttt{[2]}\\
    There was a man named Jiang, whose grandfather and father were killed by snakes when catching them. But he still made his living by catching snakes. \texttt{[1]} When Liu advised him no longer to catch snakes, the man cried and said that \hlred{\textbf{he would rather be killed by snakes than give up catching snakes.}} \texttt{[2]} Actually some villagers had already lost everything and have nothing to eat. \texttt{[3]} They could do nothing but tremble with fear when the officers went into their houses to collect taxes and struck out violently. 
    \texttt{[4]} Even dogs and chickens couldn't get any peace in such scenario, let alone humans!&{This was \hlred{\textbf{because he was able to pay taxes to the government only by catching snakes.}}}&\texttt{[2]}\\
    \midrule
        \begin{CJK}{UTF8}{gbsn}\tiny 一只狼出去找食物，偶然经过一户人家，听到小孩哭声，接着又听见老太婆的声音：“别哭了，再哭就把你扔出去喂狼。 \texttt{[1]}”狼一听心中大喜，便蹲在墙角等着，谁知等到天黑也不见把小孩扔出来。 \texttt{[2]}却又听到老太婆说：“快睡吧，别怕，狼来了，咱们就把它杀死煮了吃。 \texttt{[3]}”狼吓得一溜烟跑回了窝。 \texttt{[4]}同伴问它收获怎样，它沮丧地说：“别提了”
        \end{CJK}&\begin{CJK}{UTF8}{gbsn}\tiny 太阳落山了，狼已经等得不耐烦了.转到房前想伺机而入。\end{CJK}& \texttt{[2]}\\
    A wolf went out to look for food. It happened to pass by a \hlred{\textbf{house}}. It heard a child crying and then an old woman scared the child to say: ``Do not cry! If you cry again, I will fling out you to feed wolves right away. \texttt{[1]}'' Hearing this, the wolf was overjoyed and then squatted down and waited.  However, the child was not flung out even when \hlred{\textbf{it was dark}}. \texttt{[2]} \hlred{\textbf{Suddenly}}, the woman said: ``Don't be afraid. If the wolf comes, let's kill and eat it.''\texttt{[3]} The wolf was so frightened that he ran back to its lair. \texttt{[4]} When its friends asked it what happened, it said in dismay: ``Don't mention it.'' 
    & \hlred{\textbf{After sunset}}, the wolf was getting impatient and planned to \hlred{\textbf{break into the house}}. & \texttt{[2]}\\
	\bottomrule
    \end{tabular}
    \caption{Two SenPos examples. The special tokens from \texttt{[1]} to \texttt{[9]} refer to the candidate positions. The first/second example focuses on testing the ability to capture the inter-sentence causal/temporal relations, respectively. We highlight the entities and events implying the relations in \hlred{\textbf{red}}.}
    \label{tab:case_ssp}
\end{table*}

\paragraph{Dataset Construction} We asked annotators to refine each example so that the removed sentence has only one reasonable position in the text.
We did not allow annotators to select the first or last sentence of the original text as the removed sentence since they usually contain obvious wording features~(e.g., {``once upon a time,''} {``they lived happily together''}), which may make this task trivial. Unlike ClozeT, we allowed the texts for SenPos to be incomplete or include dialogues which also embody rich inter-sentence relations. Finally, we collected 1,663 examples for validation and testing through human annotation. And we constructed 20,000 examples automatically for training.

\subsection{Plot Completion}
We use the Plot Completion task~\cite{DBLP:conf/ijcai/Wang019b} to test the ability to make inferences based on common sense. We formulate this task as follows: given a story with a sentence removed, models should generate a sentence to complete the story and make it reasonable and coherent.

\paragraph{Dataset Construction} Prior studies~\cite{DBLP:conf/ijcai/Wang019b,paul-frank-2021-coins} automatically constructed datasets for this task based on existing datasets by randomly removing one sentence from a story. However, as shown in Table~\ref{tab:cs}, not all sentences in a story can be reasoned only based on the context and common sense. Therefore, we only used the above automatic method to construct the training data. And we adapted the ClozeT data to this task for validation and testing, since annotators have marked out the qualified sentences. 
Specifically, we randomly sampled some ClozeT examples and took the incomplete story of each example as input, and the right candidate as the target sentence to be generated. 

\subsection{Outline-conditioned Generation}
Prior works tended to test the ability of long text generation through story generation conditioned on inputs with limited information such as titles~\cite{yao2019plan}.
However, these tasks are extremely open-ended 
so that it is difficult to reliably measure the generation quality using automatic metrics~\cite{guan2020union}. 
To alleviate the issue, we introduce the Outline-conditioned Generation task~\cite{rashkin2020plotmachines}, which requires generating a coherent long-form story conditioned on an outline of characters and events. We formulate the outline as a set of out-of-order phrases, which not only narrows down the set of plausible stories but also serves for testing the controllability and planning ability of models to arrange the given events reasonably at the discourse level.



\paragraph{Dataset Construction} 
We built the dataset for this task automatically based on filtered stories. We followed~\citet{rashkin2020plotmachines} to extract the outline of a story using the RAKE algorithm~\cite{rose2010automatic}. We extract at most eight phrases for each story, and each phrase contains no more than eight words. For example, the outline for the story in Table~\ref{tab:example} is \{{``told his son with irony,''} {``purchasing flour from a mill,''} {``crossing the river,''} {``drop the sack into the river,''} {``indeed pushed the sack,''} {``familiar to his son's temper,''} {``shouted,''} {``one bag''}\}. The outline can serve as discourse-level guidance for generation models, which should rearrange the events reasonably and generate a story with a good global discourse structure, rather than focus on modeling only the local coherence.

\subsection{Overall Score}
Existing benchmarks usually summarize the performance of a model as a single score by averaging all metric scores without considering task difficulties. To encourage models to progress on those tasks where there is a more significant gap between machines and humans, we propose to 
average metric scores with different weights. Suppose that there are a total of $M$ metrics for all tasks, we derive the overall score as follows:
\begin{align}
S &= \sum_{i=1}^M\frac{w_i}{\sum_{j=1}^Mw_j}S_{i},\\
w_i&=\frac{H_i}{B_i},
\end{align}
where $H_i$, $B_i$ and $S_i$ are the score of humans, a pre-selected baseline and the evaluated model for the $i$-th metric, respectively, and $w_i$ is the weight for this metric. Intuitively, the metric scores where the baseline model has a larger gap with humans will have a larger weight when computing the overall score. We use BERT and GPT2 as the baseline models for the understanding and generation tasks in LOT, respectively. 

\section{Long Text Pretraining Model}
To provide more flexibility on both understanding and generation tasks, we build LongLM following the original encoder-decoder design of Transformer~\cite{vaswani2017attention} with three different sizes, as shown in Table~\ref{tab:modelset}. We follow \citet{cui2020revisiting} to use a sentencepiece vocabulary of 32,000 wordpieces~\cite{kudo2018sentencepiece}. And we set the maximum sequence length to 512 for both the encoder and decoder.

\begin{table}[!ht]
\scriptsize
    \centering
    \begin{tabular}{l|ccccccc}    
    \toprule
\textbf{Versions}&$d_{\rm m}$&$d_{\rm ff}$&$d_{\rm kv}$&$n_{\rm h}$&$n_{\rm e}$/$n_{\rm d}$&{\# P}\\
\midrule
\textbf{Small}&512&2,048&64&8&6/6&60M\\
\textbf{Base}&768&3,072&64&12&12/12&223M\\
\textbf{Large}&1,536&3,072&64&12&24/32&1B\\
\bottomrule
    \end{tabular}
    \caption{Hyper-parameter settings for different versions of LongLM. $d_{\rm m}$, $d_{\rm ff}$ and $d_{\rm kv}$ are the dimension of hidden states, the feed forward layers, the keys/values in the self-attention layers, respectively. $n_{\rm h}$ is the number of attention heads. $n_{\rm e}$ and $n_{\rm d}$ denote the number of hidden layers for the encoder and decoder, respectively. \# P is the number of parameters.}
    \label{tab:modelset}
\end{table}

\paragraph{Pretraining Data} We collect 120G novels as the pretraining data for LongLM, which cover various topics such as romance, military, etc. Since a novel is usually much longer than the maximum input and output length of LongLM, we split a novel into multiple segments for pretraining.

\begin{figure}[!ht]
\includegraphics[width=\linewidth]{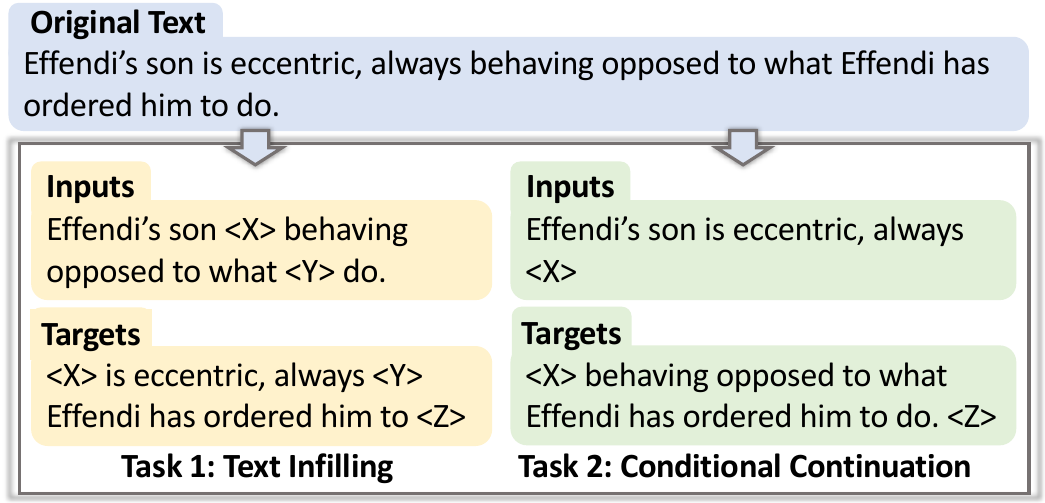}
  \caption{Schematic of the pretraining tasks. <X> and <Y> is the special tokens used for masking spans. <Z> is the ``end of sequence'' token.}
  \label{tab:pretrain_task}
\end{figure}

\paragraph{Pretraining Tasks} Encoder-decoder models are trained typically by maximizing the likelihood of the target output given an input. To improve capacities of both the encoder and decoder, 
we propose to train LongLM with two pretraining tasks including text infilling~\cite{raffel2020exploring} and conditional continuation~\cite{radford2019language}. For the first task, the input is a text where a number of spans are sampled and replaced by special tokens with unique IDs,
while the output is the spans delimited by the special tokens used in the input. The lengths of masked spans are drawn from a Poisson distribution with $\lambda$=3 and all masked tokens compress 15\% of the original texts. As for the second task, the input and output are respectively the front and back half of a text, which is split into two parts randomly. We show an example of the pretraining tasks in Figure~\ref{tab:pretrain_task}.

\paragraph{Pretraing Details} We set the learning rate to 1e-4 with the Adam optimizer 
and the batch size to 1,000. We pretrained LongLM for 2.5M steps. It took about two months to train the largest model using eight NVIDIA V100 GPUs.

\paragraph{Model Performance} To assess the performance of LongLM on the pretraining tasks, we randomly separated out 1,000 texts from the initial pretraining data for testing, which were never seen in the pretraining phase. We used perplexity and BLEU-$n$~($n$=3,4) to evaluate both pretraining tasks. And we generated outputs using the greedy decoding algorithm for the text infilling task, and top-$k$ sampling~\cite{fan2018hierarchical} with $k=40$ and a softmax temperature of 0.7~\cite{goodfellow2014generative} for the conditional continuation task. As shown in Table~\ref{tab:longlm_performance}, the performance improves substantially as the number of parameters increases.

\begin{table}[!ht]
\scriptsize
    \centering
    \begin{tabular}{l|cc|cc}    
    \toprule
\multirow{2}{*}{\textbf{Models}}&\multicolumn{2}{c|}{\textbf{TextInfill}}&\multicolumn{2}{c}{\textbf{CondCont}}\\
&\textbf{PPL}&\textbf{BLEU-3/4}&\textbf{PPL}&\textbf{BLEU-3/4}\\
\midrule
\textbf{LongLM$_{\rm small}$}&11.61&73.80/68.96&22.91&5.30/2.43\\
\textbf{LongLM$_{\rm base}$}& \underline{8.24}&\underline{75.65}/\underline{71.05}&\underline{17.03}&\underline{5.73}/\underline{2.64}\\
\textbf{LongLM$_{\rm large}$}&\textbf{6.50}&\textbf{77.08}/\textbf{72.65}&\textbf{14.08}&\textbf{8.91}/\textbf{5.97}\\
\bottomrule
    \end{tabular}
    \caption{Perplexity (PPL) and BLEU scores of LongLM for text infilling~(TextInfill) and conditional continuation~(CondCont). The best performance is in \textbf{bold} and the second best is \underline{underlined}.}
    \label{tab:longlm_performance}
\end{table}

\section{Experiments}
In this section, we tested LongLM and existing models on LOT with automatic and manual evaluation. Furthermore, we conducted extensive experiments to investigate the potential biases of the ClozeT and SenPos datasets~(Section \ref{bias_invest}), and measure the overlap between training and testing data~(Section~\ref{memory_invest}).
\subsection{Evaluated Models}
We evaluated the following models, which are implemented based on the register models of HuggingFace Transformers\footnote{\url{https://huggingface.co/models}}: \textbf{(1) Vanilla Transformer:} It has the same architecture as BERT$_{\rm base}$ except that the number of layers is set to 3~\cite{vaswani2017attention}. 
\textbf{(2) BERT:} It's implemented based on the \textit{bert-base-Chinese} register model~\cite{devlin2018bert}. \textbf{(3) RoBERTa:} It's implemented based on the \textit{hfl/chinese-roberta-wwm-ext} register model~\cite{cui2020revisiting}. \textbf{(4) GPT2:} It's implemented based on the \textit{uer/gpt2-chinese-cluecorpussmall} register model~\cite{zhao2019uer}. \textbf{(5) mT5:} It's implemented based on the \textit{google/mt5-base} register model~\cite{xue2021mt5}. We set all the baseline models to the base version due to limited computational resources.

To show the generic benefits of the pretraining data of LongLM for long text modeling, we pretrained a left-to-right language model from scratch on the data with the standard language modeling objective. This model has the same architecture as GPT2$_{\rm base}$ and is denoted as \textbf{GPT2$^\dagger_{\rm base}$}. Moreover, we evaluated two task-specific pretraining models including PlotMachines~(\textbf{PM})~\cite{rashkin2020plotmachines} and {Plan\&Write}~(\textbf{PW})~\cite{yao2019plan}, and two typical non-pretrained models including \textbf{ConvS2S}~\cite{gehring2017convolutional} and \textbf{Fusion}~\cite{fan2018hierarchical} on the generation tasks in LOT. 
We used GPT2$_{\rm base}$ as the backbone model of PM and PW. For PM, we regard input sentences~(for PlotCom) or input phrases~(for OutGen) as the plot elements used in the memory network, and update the memory representations at each step of decoding. As for PW, we take a keyword extracted from the target sentence using the RAKE algorithm~(for PlotCom) or the sorted input phrases in order~(for OutGen) as the intermediate representations for planning. 
We implemented these models based on the codes provided by the original papers. 

\subsection{Experiment Settings}
\paragraph{Understanding Tasks} For both tasks, 
we encode the input of each example and then predict a distribution over all candidates by normalizing the dot-product values between the representations of each candidate and the context. We use the candidate with the maximum probability as the prediction result. For ClozeT, we represent a candidate using the hidden state at the end of it, and we regard the hidden state at the position of the removed sentence appearing in the original text as the context representation. And for SenPos, we take the hidden state at each candidate position as the candidate representation and the hidden state at the end of the removed sentence as the context representation.
When evaluating mT5 and LongLM, we feed the same input into the encoder and decoder ~\cite{bart} and use the hidden states of the decoder for prediction in the above way.

\paragraph{Generation Tasks} For PlotCom, we take the incomplete story of an example as input to generate the missing sentence. And for OutGen, we concatenate all phrases in an outline with special tokens as input to generate a story. 

\paragraph{Hyper-Parameters}
For all models, we set the batch size to 12, the maximum sequence length to 512, and the learning rate to 3e-5. We decode outputs use top-$k$ sampling 
with $k=40$ and a softmax temperature of 0.7 
for the generation tasks.

\begin{table}[!t]
\scriptsize
    \centering
    \begin{tabular}{l|c|cc|c}
    \toprule
\textbf{Models}&\textbf{\# P}&\textbf{ClozeT}&\textbf{SenPos}&\textbf{Overall}\\
\midrule
\midrule
\multicolumn{5}{c}{\textbf{Validation Set}}\\
\midrule
\textbf{Transformer}&38M&55.78& 17.38&31.46\\
\midrule
\textbf{BERT$_{\rm base}$} &102M & {70.75} & 40.13 & 51.36\\
\textbf{RoBERTa$_{\rm base}$}&102M & {72.11} & {51.63} & 59.14 \\
\textbf{GPT2$_{\rm base}$}&102M & 70.07 & 37.78& 49.62\\ 
\textbf{GPT2$_{\rm base}^\dagger$}&102M&74.49&39.25&52.17\\
\textbf{mT5$_{\rm base}$}&582M&72.45&{63.25}&66.62\\
\midrule
\textbf{LongLM$_{\rm small}$}&60M&{73.81}&48.75 & 57.94\\
\textbf{LongLM$_{\rm base}$}&223M&\underline{75.17}&\underline{64.38} &\underline{68.34} \\
\textbf{LongLM$_{\rm large}$}&1B&\textbf{79.93}&\textbf{70.00}&\textbf{73.64}\\
\midrule
\textit{\textbf{Humans}}&\textit{N/A}&\textit{99.00}&\textit{97.00}&\textit{97.73}\\
\midrule
\textbf{$w_i$}&\textit{N/A}&\textit{0.37}&\textit{0.63}&\textit{1.00}\\
\midrule
\midrule
\multicolumn{5}{c}{\textbf{Test Set}}\\
\midrule
\textbf{Transformer}&38M&54.42&16.34 & 31.23\\
\midrule
\textbf{BERT$_{\rm base}$} & 102M&69.39 &43.68 &53.74\\ 
\textbf{RoBERTa$_{\rm base}$} & 102M &67.69 & {51.35} & 57.74\\
\textbf{GPT2$_{\rm base}$}&102M&73.13&37.25&51.28\\
\textbf{GPT2$_{\rm base}^\dagger$}&102M&76.87&39.28&53.98\\
\textbf{mT5$_{\rm base}$}&582M&75.17&{61.41}&{66.79}\\
\midrule
\textbf{LongLM$_{\rm small}$}&60M&{77.21}&53.07 & 62.51\\
\textbf{LongLM$_{\rm base}$}&223M&\underline{77.55}&\underline{62.34} &\underline{68.29}\\
\textbf{LongLM$_{\rm large}$}&1B&\textbf{80.61}&\textbf{69.41}&\textbf{73.39}\\
\midrule
\textit{\textbf{Humans}}&\textit{N/A}&\textit{100.00}&\textit{98.00}&\textit{98.78}\\
\midrule
\textbf{$w_i$}&\textit{N/A}&\textit{0.39}&\textit{0.61}&\textit{1.00}\\
\bottomrule
    \end{tabular}
    \caption{Accuracy~(\%) on the understanding tasks in LOT. \# P means the number of parameters. The best performance is in \textbf{bold} and the second best is \underline{underlined}. $w_i$ is the metric weight with BERT as the baseline model when computing the overall score.}
    \label{tab:unLOT}
\end{table}

\begin{table*}[!t]
\scriptsize
    \centering
    \begin{tabular}{l|c|cccc|cccccc|c}
    \toprule
\multirow{2}{*}{\textbf{Models}}&\multirow{2}{*}{\textbf{\# P}}&\multicolumn{4}{c|}{\textbf{PlotCom}}&\multicolumn{6}{c|}{\textbf{OutGen}}&\multirow{2}{*}{\textbf{Overall}}\\
&&\textbf{B-1}&\textbf{B-2}&\textbf{D-1}&\textbf{D-2}&\textbf{B-1}&\textbf{B-2}&\textbf{D-1}&\textbf{D-2}&\textbf{Cover}&\textbf{Order}\\
\midrule
\midrule
\multicolumn{13}{c}{\textbf{Validation Set}}\\
\midrule
\textbf{ConvS2S}& 58M&18.92&4.18&6.31&32.18&29.23&10.38&3.45&21.79&14.81&25.34 & 11.85\\
\textbf{Fusion}&109M&20.56&4.69&8.63&35.73&29.22&10.34&3.39&22.67&17.41&26.55& 12.61\\
\midrule
\textbf{GPT2$_{\rm base}$}&102M&{22.67}&6.22&{24.75}&{70.57}&30.43&14.87&10.95&{44.38}&60.90&55.52 & 20.24\\
\textbf{GPT2$_{\rm base}^\dagger$}&102M&22.49&5.43&\textbf{26.88}&\textbf{74.87}&35.29&18.31&13.89&51.36&64.01&57.64&21.73\\
\textbf{{PM}}&102M&22.11&5.49&23.89&69.74&31.81&14.94&12.99&50.56&62.98&56.75&20.45\\
\textbf{PW}&102M&22.45&5.57&25.64&71.54&35.84&18.47&11.86&47.62&64.93&57.30&21.48\\
\textbf{mT5$_{\rm base}$}&582M&22.56&6.46&{24.44}&{71.31}&{36.71}&{22.25}&{14.52}&{50.01}&{77.98}&\underline{63.15}&{23.53}\\
\midrule
\textbf{LongLM$_{\rm small}$}&60M&21.78&{7.11}&20.17&59.63&{35.03}&{19.17}&{10.80}&39.70&{62.53}&{56.53} & {21.02}
\\
\textbf{LongLM$_{\rm base}$}&223M&\underline{22.91}&\underline{8.28}&{22.16}&{63.54}&\underline{40.33}&\underline{24.29}&\underline{14.66}&\underline{51.82}&\underline{79.60}&{62.78} & \underline{24.75}\\
\textbf{LongLM$_{\rm large}$}&1B&\textbf{23.76}&\textbf{8.70}&{\underline{25.93}}&{\underline{72.18}}&\textbf{42.79}&\textbf{24.91}&\textbf{16.13}&\textbf{57.71}&\textbf{80.46}&\textbf{64.36}&\textbf{26.12}\\
\midrule
\textit{\textbf{Truth}}&\textit{N/A}&\textit{100.00}&\textit{100.00}&\textit{35.32}&\textit{84.33}&\textit{100.00}&\textit{100.00}&\textit{21.66}&\textit{71.43}&\textit{100.00}&\textit{100.00} & \textit{92.23}\\
\midrule
\textbf{$w_i$}&\textit{N/A}&\textit{0.11}&\textit{0.40}&\textit{0.04}&\textit{0.03}&\textit{0.08}&\textit{0.17}&\textit{0.05}&\textit{0.04}&\textit{0.04}&\textit{0.04}&\textit{1.00}\\
\midrule
\midrule
\multicolumn{13}{c}{\textbf{Test Set}}\\
\midrule
\textbf{ConvS2S}& 58M&19.60&4.20&6.00&32.42&29.00&10.14&1.60&13.95&15.45&25.77 & 11.27\\
\textbf{Fusion}&109M&20.52&4.90&8.43&35.09&28.77&10.22&1.47&14.12&17.10&26.36&11.91\\
\midrule
\textbf{GPT2$_{\rm base}$}&102M&{22.94}&5.76&{24.69}&{70.30}&30.17&14.91&7.62&{36.87}&60.87&55.90&19.21\\
\textbf{GPT2$_{\rm base}^\dagger$}&102M&22.45&5.38&\textbf{26.08}&\textbf{73.26}&35.79&18.68&9.89&43.52&64.43&56.96&20.76\\
\textbf{{PM}}&102M&22.87&5.75&24.08&\underline{71.19}&31.85&15.24&8.62&41.32&63.15&57.21&19.77\\
\textbf{\textbf{PW}}&102M&22.76&6.07&25.55&70.72&35.12&17.96&8.68&40.17&63.70&55.17&20.52\\
\textbf{mT5$_{\rm base}$}&582M&22.52&6.48&{24.33}&{70.53}&{36.33}&{22.07}&\underline{10.90}&{43.65}&{78.66}&\underline{63.79}&{22.59}\\
\midrule
\textbf{LongLM$_{\rm small}$}&60M&22.05&{7.45}&19.93&59.79&{34.48}&{19.17}&{7.93}&{34.25}&{63.75}&{57.64} & {20.48}\\
\textbf{LongLM$_{\rm base}$}&223M&\underline{23.28}&\underline{8.58}&{21.37}&{62.43}&\underline{40.25}&\underline{24.15}&{10.75}&\underline{44.40}&\underline{79.88}&{63.67} & \underline{23.93}\\
\textbf{LongLM$_{\rm large}$}&1B&\textbf{24.20}&\textbf{9.06}&\underline{25.75}&{71.08}&\textbf{42.10}&\textbf{24.77}&\textbf{12.04}&\textbf{50.29}&\textbf{81.48}&\textbf{64.82}&\textbf{25.29}\\
\midrule
\textit{\textbf{Truth}}&\textit{N/A}&\textit{100.00}&\textit{100.00}&\textit{35.01}&\textit{84.56}&\textit{100.00}&\textit{100.00}&\textit{15.71}&\textit{63.46}&\textit{100.00}&\textit{100.00} & \textit{91.64}\\
\midrule
\textbf{$w_i$}&\textit{N/A}&\textit{0.10}&\textit{0.42}&\textit{0.03}&\textit{0.03}&\textit{0.08}&\textit{0.16}&\textit{0.05}&\textit{0.04}&\textit{0.04}&\textit{0.04}&\textit{1.00}\\
\bottomrule
    \end{tabular}
    \caption{Evaluation results on the generation tasks in LOT. \# P means the number of parameters. The best performance is in \textbf{bold} and the second best is \underline{underlined}. $w_i$ is the metric weight with GPT2$_{\rm base}$ as the baseline model when computing the overall score.}
    \label{tab:genLOT}
\end{table*}

\subsection{Automatic Evaluation}
\paragraph{Metrics}~{}\label{sec:metrics}
We use accuracy to evaluate the understanding tasks. As for generation tasks, we use BLEU-$n$ (B-$n$) and Distinct-$n$ (D-$n$) to evaluate the $n$-gram overlap with ground-truth texts~\cite{papineni2002bleu} and $n$-gram generation diversity~\cite{li2016diversity}, respectively. We set $n=1,2$ for both generation tasks. Additionally, we also use the following two metrics to evaluate OutGen: \textbf{(1) Coverage~(Cover):} It is used to evaluate the generation controllability, which is computed as the average Rouge-L recall score~\cite{lin-2004-rouge} between the generated text and each input phrase. 
A higher coverage score indicates the generated text covers more input phrases. \textbf{(2) Order:} It is used to measure the gap between the positional orders of input phrases appearing in the generated texts and ground-truth texts.
Specifically, we compute the order score as the average ratio of the number of inversions in the generated story to the number of all position pairs of any two phrases. An inversion refers to a position pair that are out of the ground-truth order. And we use the position of the longest common subsequence between a story and a phrase as the position of the phrase in the story. Because an input phrase does not always appear in the generated story, we regard all  position pairs of such a phrase and others as inversions.

\begin{figure}[!t]
\includegraphics[width=\linewidth]{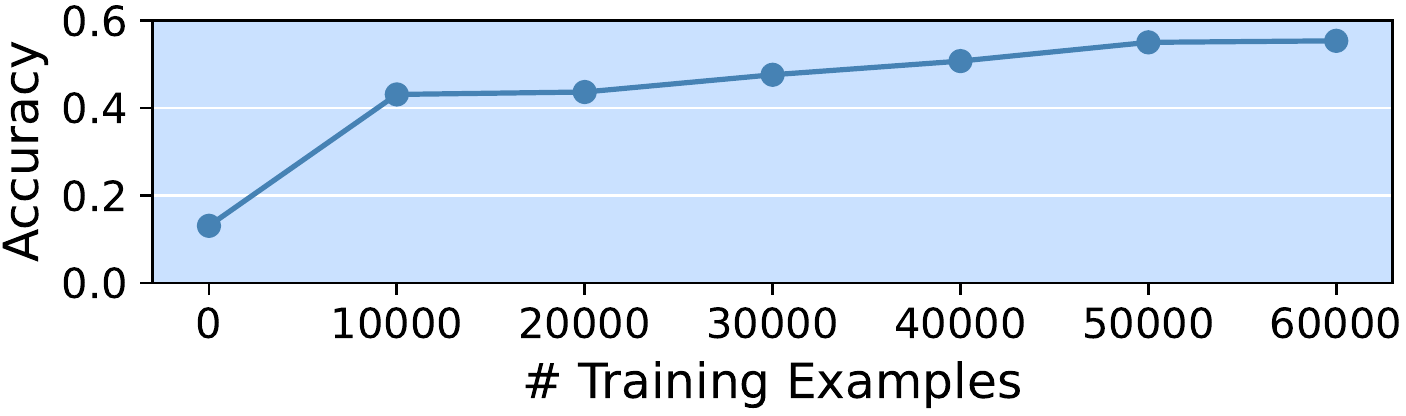}
  \caption{Accuracy of BERT for SenPos as  the size of training data increases.}
  \label{tab:size}
\end{figure}

\paragraph{Results}~{}
Table~\ref{tab:unLOT} and \ref{tab:genLOT} show the results on the understanding and generation tasks, respectively. 
To obtain the human performance on the understanding tasks, we randomly sampled 100 examples from the validation set or test set and hired three crowd-sourced annotators~(native Chinese speakers) to do these tasks. We made final decisions among them through majority voting. All results show an almost perfect inter-annotator agreement with Fleiss's $\kappa>0.85$~\cite{Fleiss1971Measuring}. For generation tasks, we regard the scores of ground-truth texts as human performance. 

We summarize the evaluation results as follows: \textbf{(1)} Pretrained models have significantly better performance than non-pretrained models. \textbf{(2)} LongLM$_{\rm large}$ outperforms other baselines substantially on both the understanding and generation tasks. 
LongLM$_{\rm base}$/LongLM$_{\rm small}$ achieves better overall scores with half fewer parameters than mT5/GPT2. 
\textbf{(3)} By comparing GPT2$^\dagger$ and GPT2, we can derive that our pretraining data can effectively improve the ability to model long texts. \textbf{(4)} LongLM$_{\rm small}$ has a better performance than GPT2$^\dagger$ on the understanding tasks, and is comparable with GPT2$^\dagger$ on the generation tasks,  suggesting the benefits of the encoder-decoder framework and the text infilling task. \textbf{(5)} It is still extremely challenging for all models to capture the commonsense and inter-sentence discourse relations between events in long texts for tackling the ClozeT and SenPos tasks. 
Furthermore, we investigate how the size of training data influences the accuracy of BERT for SenPos. The result in Figure~\ref{tab:size} indicates the necessity to develop better representations of discourse relations instead of relying only on increasing the data size. 
\textbf{(6)} The results on the generation tasks show that LongLM does well in generating more word overlaps with references than similar-sized baselines for both tasks, and covers more input phrases and arranges them in correct orders for OutGen. But LongLM underperforms GPT2-based models in terms of diversity on PlotCom. \textbf{(7)} Dynamically tracking plot states (i.e., PM) does not bring significant improvement on the generation tasks compared with GPT2, suggesting that it may require modeling the discourse structure explicitly to tackle the generation tasks. And the superiority of PW to GPT2 on OutGen further indicates the benefit of modeling discourse-level features. 
In summary, we believe LOT will serve as an effective evaluation for capturing the commonsense and discourse relations of long texts beyond the surface events, and generating coherent and controllable long-form texts.


\subsection{Manual Evaluation}
Since automatic metrics may be unreliable for evaluating NLG~\cite{guan2020union}, we conducted a point-wise manual evaluation to measure the disparity between machines and humans for the generation tasks in LOT. For each task, we randomly sampled 100 examples from the test set and obtained 100 ground-truth texts and 300 generated texts from three typical models including GPT2$_{\rm base}$, mT5$_{\rm base}$ and LongLM$_{\rm large}$. For each text along with the input, we hired three crowd-sourced workers 
to judge its quality with a binary score~(1 for good, and 0 otherwise) in terms of three aspects: (1) \textit{grammaticality}~(intra-sentence grammar quality of generated texts), (2) \textit{coherence}~(causal and temporal dependencies within generated texts), and (3) \textit{relatedness to inputs}~(reasonable logical connections to the input context for PlotCom; and reasonable utilization of input phrases for OutGen). These aspects are independently evaluated. We made final decisions among three annotators through majority voting. We show the annotation instructions in the appendix.

\begin{table}[!t]
\scriptsize
    \centering
    \begin{tabular}{l|c|c|c}
    \toprule
    \textbf{Models}&\textbf{Gram ($\kappa$)}&\textbf{Cohe~($\kappa$)}&\textbf{Relat~($\kappa$)}\\
    \midrule
    \midrule
    \multicolumn{4}{c}{\textbf{Task: PlotCom}}\\
    \midrule
    \textbf{GPT2$_{\rm base}$}&0.84~(0.49)&0.41~(0.71)&0.01~(0.50)\\
    \textbf{mT5$_{\rm base}$}&0.85~(0.24)&0.53~(0.65)&0.01~(0.50)\\
    \textbf{LongLM$_{\rm large}$}&\textbf{0.95}~(0.48)&\textbf{0.82}~(0.64)&\textbf{0.09}~(0.69)\\
    \textit{\textbf{Truth}}&\textit{1.00}~(1.00)&\textit{1.00}~(1.00)&\textit{0.99}~(0.49)\\
    \midrule
    \midrule
    \multicolumn{4}{c}{\textbf{Task: OutGen}}\\
    \midrule
    \textbf{GPT2$_{\rm base}$}&0.54~(0.52)&0.18~(0.52)&0.39~(0.43)\\
    \textbf{mT5$_{\rm base}$}&0.53~(0.26)&0.08~(0.46)&0.49~(0.38)\\
    \textbf{LongLM$_{\rm large}$}&\textbf{0.81}~(0.23)&\textbf{0.37}~(0.43)&\textbf{0.62}~(0.45)\\
    \textit{\textbf{Truth}}&\textit{1.00}~(1.00)&\textit{1.00}~(1.00)&\textit{1.00}~(1.00)\\   
    \bottomrule
    \end{tabular}
    \caption{Manual evaluation results for PlotCom and OutGen in terms of grammaticality~(\textbf{Gram}), coherence~(\textbf{Cohe}) and relatedness~(\textbf{Relat}). The best performance is highlighted in \textbf{bold}. All results show a fair inter-annotator agreement with Fleiss' $\kappa>0.2$.}
    \label{tab:man_eva}
\end{table}

Table~\ref{tab:man_eva} shows the evaluation results. 
For both tasks, LongLM outperforms GPT2 and mT5 significantly in all aspects~($p<0.05$, sign test). However, it is difficult for all models to generate a logical completion for PlotCom ~(relatedness score $<0.1$), showing their poor ability to capture commonsense and inter-sentence relations. And the big gap between LongLM and humans also proves both tasks challenging to existing generation models. 
We also observe the positive correlation between the manual evaluation and automatic evaluation~(Table~\ref{tab:genLOT}), suggesting that it may be acceptable to use automatic evaluation to compare and improve models on the generation tasks in LOT.

\subsection{Bias Investigation}\label{bias_invest}

It is essential to investigate potential biases of a dataset, which may leak information about target labels and enable models to easily use shortcuts to handle complex inputs without actually mastering the focused abilities~\cite{ribeiro-etal-2020-beyond}. 
Therefore, we experimented with the following baselines to inspect the ClozeT and SenPos datasets: \textbf{(1) Random:} It chooses a candidate randomly. \textbf{(2) Majority:} It chooses the candidate with an index that is most frequently selected in the training set. \textbf{(3) Length:} For ClozeT, it chooses the candidate that contains more words; And for SenPos, it chooses the position of which the adjacent sentences have the closest number of words to the removed sentence. 
\textbf{(4) BLEU-$n$:} For ClozeT, it chooses the candidate with a higher BLEU-$n$ score~\cite{papineni2002bleu}  with the context; And for SenPos, it chooses the position of which the adjacent sentences have the largest average BLEU-$n$ score with the removed sentence~($n$=1,2).
\textbf{{(5) Sentiment:}} For ClozeT, it chooses the candidate with a higher sentiment score computed by an off-the-shelf Chinese sentiment analyzer\footnote{\url{https://github.com/isnowfy/snownlp}}; And for SenPos, it chooses the position where the average sentiment score of its adjacent two sentences is the closest to the score of the removed sentence. 
\textbf{(6) Discourse Markers:} For ClozeT, it chooses the candidate where its adjacent sentences contain a discourse marker matching with it. For example, if ``because'' 
occurs in the last sentence before the position of the candidates, this baseline will choose the candidate that contains ``so''
\footnote{Different from English, paired discourse markers like ``because''-``so'' should be used together in Chinese.}. If there does not exist such paired markers in an example or there are multiple eligible candidates,
this baseline will 
randomly choose one. 
The setting of this baseline for SenPos is similar to ClozeT. We manually define 24 marker pairs for this baseline. \textbf{(7) BERT w/o Context:} We fine-tuned BERT to directly choose without taking the context as input~\cite{schwartz2017effect}. \textbf{(8) BERT w/o Long:} It is used to study whether solving these tasks requires modeling long-range dependencies. For ClozeT, we fine-tuned BERT to choose with only the adjacent sentences of the removed sentence as input. And for SenPos, we encoded each position and its adjacent sentences respectively using BERT and then took the hidden states at these positions for prediction. 
These baselines cover different levels of features ranging from the token level~(e.g., \textit{Length}), the sentence level~(e.g., \textit{Sentiment}) to the discourse level~(e.g., \textit{Discourse Markers}, \textit{BERT w/o Context}). We believe that these baselines will provide a comprehensive inspection for the potential biases of our datasets.


\begin{table}[!t]
\footnotesize
    \centering
    \begin{tabular}{l|cc}
    \toprule
\textbf{Baselines}&\textbf{ClozeT}&\textbf{SenPos}\\
\midrule
\textbf{Random}&50.00&16.03\\
\textbf{Majority}&52.72&16.24\\
\textbf{Length}&52.72&16.45\\
\textbf{BLEU-1/2}&46.94/48.98&14.14/14.95\\
\textbf{Sentiment}&50.34&16.49\\
\textbf{Discouse Markers}&45.92&9.15\\
\textbf{BERT w/o Context}&57.82&18.08\\
\textbf{BERT w/o Long}&62.24&19.00\\
\midrule
\textit{\textbf{BERT}}&\textit{69.39}&\textit{43.68}\\
\bottomrule
    \end{tabular}
    \caption{Accuracy~(\%) of different baselines on the test sets of ClozeT and SenPos for bias investigation. We use the results of BERT as a reference.}
    \label{tab:diff}
\end{table}

As shown in Table~\ref{tab:diff}, 
both tasks can not be trivially solved by these baselines, suggesting that the datasets may be free of biases in terms of the above features. Therefore, we believe that the tasks can focus on testing the ability of models to capture long-range commonsense and discourse relations.

\begin{table}[!t]
\scriptsize
    \centering
    \begin{tabular}{l|cccc}
    \toprule
\textbf{Tasks}&\textbf{ClozeT}&\textbf{SenPos}&\textbf{PlotCom}&\textbf{OutGen}\\
\midrule
\midrule
\multicolumn{5}{c}{\textbf{Overlap with the Training Sets}}\\
\midrule
\textbf{Percent}&0.00\%&0.62\%&0.02\%&0.00\%\\
\midrule
\# \textbf{8-grams}&0&1,040&6&2\\
\# \textbf{Exam}&0&45&3&2\\
\# \textbf{Exam$_{>10\%}$}&0&17&0&0\\
\textbf{Max Percent}&0.00\%&60.98\%&2.53\%&1.00\%\\
\midrule
\midrule
\multicolumn{5}{c}{\textbf{Overlap with the Pretraining Data}}\\
\midrule
\textbf{Percent}&0.67\%&4.68\%&0.38\%&1.22\%\\
\midrule
\# \textbf{8-grams}&172&7,844&151&1,212\\
\# \textbf{Exam}&83&486&88&161\\
\# \textbf{Exam$_{>10\%}$}&4&71&1&26\\
\textbf{Max Percent}&47.22\%&60.96\%&30.77\%&41.18\%\\
\bottomrule
    \end{tabular}
    \caption{Overlapping analysis for the test sets of the four tasks with respect to their own training sets or the pretraining data of LongLM. We compute the following statistics: \textbf{(1) Percent}: the percentage of 8-grams from the test set that are also in the training sets or the pretraining data; \textbf{(2)} \# \textbf{8-grams}: the number of overlapped 8-grams; \textbf{(3) }\# \textbf{Exam}: the number of examples that contain at least one overlapped 8-gram; \textbf{(4)} \# \textbf{Exam$_{>10\%}$}: the number of examples that have more than 10\% overlapped 8-grams. \textbf{(4) Max Percent}: the maximum percentage of overlapped 8-grams from an example.}
    \label{tab:overlap}
\end{table}

\subsection{Memorization Investigation}\label{memory_invest}
Overlap between training and test data may result in an over-reporting of the generalization performance of machines. Therefore, it is necessary to investigate how many test data also show up in the training data. To this end, we follow \citet{radford2019language} to measure the overlap between two datasets by calculating the percentage of 8-grams from one that are also in the other. We use the \texttt{jieba} tokenizer for tokenization. 

Table~\ref{tab:overlap} shows the overlapping analysis for test sets of the four tasks in LOT.  
We can see that all test sets have less than 1\% overlap with their own training sets. Notably, there are 17 test examples of SenPos that contain more than 10\% overlapped 8-grams with the training set. This is because a training example and a test example may come from the same story, and thus they share similar information~(e.g., characters, locations). A test example contains at most 60.98\% overlapped 8-grams, suggesting that the training set and test set do not include exactly the same example. As for the pretraining data of LongLM, the test sets of ClozeT and PlotCom still have less than 1\% overlap. However, there are dozens of test examples in SenPos and OutGen that contain more than 10\% overlapped 8-grams. Through manual inspection of the overlaps, we found that they mainly come from idioms, proverbs and classic fairy tales, which may be part of some novels in the pretraining data.

\begin{table}[!t]
\scriptsize
    \centering
    \begin{tabular}{l|cc|c}
    \toprule
    \multirow{2}{*}{\textbf{SenPos}}&
    \multirow{2}{*}{\textbf{Total}}&\textbf{w/o Overlap}&\multirow{2}{*}{\textbf{$\Delta$}}\\
    &&\textbf{(Training Set)}&\\
    \midrule
    \# \textbf{Exam}&863&846&N/A\\
    \midrule
    \textbf{mT5$_{\rm base}$}&61.41\%&61.82\%&+0.41\%\\
    \textbf{LongLM$_{\rm large}$}&69.41\%&69.50\%&+0.09\%\\
    \midrule
    \midrule
    \multirow{2}{*}{\textbf{SenPos}}&
    \multirow{2}{*}{\textbf{Total}}&\textbf{w/o Overlap}&\multirow{2}{*}{\textbf{$\Delta$}}\\
    &&\textbf{(Pretraining Data)}&\\
    \midrule
    \# \textbf{Exam}&863&792&N/A\\
    \midrule
    \textbf{mT5$_{\rm base}$}&61.41\%&61.24\%&-0.17\%\\
    \textbf{LongLM$_{\rm large}$}&69.41\%&69.32\%&-0.09\%\\
\bottomrule
    \end{tabular}
    \caption{Accuracy on the test set of SenPos. \textbf{Total} means using the whole test set while \textbf{w/o Overlap} means excluding the examples that have more than 10\% overlapped 8-grams with the training set or pretraining data from the test set. \# \textbf{Exam} is the number of examples. \textbf{$\Delta$} denotes the change of accuracy when excluding the overlapping data compared with using the total test set.}
    \label{tab:spp_acc_overlap}
\end{table}

\begin{table}[!t]
\scriptsize
    \centering
    \begin{tabular}{l|cc|c}
    \toprule
    \multirow{2}{*}{\textbf{OutGen}}&
    \multirow{2}{*}{\textbf{Total}}&\textbf{w/o Overlap}&\multirow{2}{*}{\textbf{$\Delta$}}\\
    &&\textbf{(Pretraining Data)}&\\
    \midrule
    \# \textbf{Exam}&729&703&N/A\\
    \midrule
    \textbf{mT5$_{\rm base}$}&36.33&36.45&+0.12\\
    \textbf{LongLM$_{\rm large}$}&42.10&42.22&+0.12\\
\bottomrule
    \end{tabular}
    \caption{BLEU-1 score on the test set of OutGen. Other notations are the same as Table~\ref{tab:spp_acc_overlap}.}
    \label{tab:outgen_bleu_overlap}
\end{table}

To investigate how the overlapping data influence the measurement of models' performance, we re-evaluated LongLM$_{\rm large}$ on the test sets of SenPos and OutGen with exclusion of the examples that have more than 10\% overlapped 8-grams with the training sets or pretraining data. We also used mT5$_{\rm base}$ as a baseline in the same setting of LongLM. The results for SenPos and OutGen are shown in Table~\ref{tab:spp_acc_overlap} and Table~\ref{tab:outgen_bleu_overlap}, respectively. The change of accuracy or BLEU-1 score is very marginal for both mT5 and LongLM when excluding the overlapping data, suggesting that the superior performance of LongLM is rarely attributable to the memorization of training data. Therefore, we believe that it is fair to compare LongLM and other models on these tasks.

\section{Conclusions}
We present LOT, a story-centric benchmark for Chinese long text understanding and generation. LOT includes two story understanding tasks and two story generation tasks, which comprehensively investigate the abilities of commonsense reasoning, controllable generation, and modeling inter-sentence relations and the global discourse structures. We provide standard datasets for the four tasks, which are constructed based on human-written stories processed by automatic and manual annotation. Furthermore, we release a new Chinese long text pretraining model LongLM, 
which outperforms strong baseline models substantially on both the understanding and generation tasks in LOT. The LOT benchmark, the pretraining model, and the evaluation platform will encourage further research on Chinese long text modeling.

\section{Acknowledgement}
This work was supported by the National Science Foundation for Distinguished Young Scholars (with No. 62125604) and the NSFC projects (Key project with No. 61936010 and regular project with No. 61876096). This work was also supported by the Guoqiang Institute of Tsinghua University, with Grant No. 2019GQG1 and 2020GQG0005. We would also like to thank our action editor, Dipanjan Das, and the anonymous reviewers for their invaluable suggestions and feedback.

\bibliography{tacl2018}

\begin{thebibliography}{58}
\expandafter\ifx\csname natexlab\endcsname\relax\def\natexlab#1{#1}\fi

\bibitem[{Agarwal et~al.(2013)Agarwal, Kotalwar, and
  Rambow}]{agarwal2013automatic}
Apoorv Agarwal, Anup Kotalwar, and Owen Rambow. 2013.
\newblock Automatic extraction of social networks from literary text: A case
  study on alice in wonderland.
\newblock In \emph{Proceedings of the Sixth International Joint Conference on
  Natural Language Processing}, pages 1202--1208.

\bibitem[{Akoury et~al.(2020)Akoury, Wang, Whiting, Hood, Peng, and
  Iyyer}]{akoury2020storium}
Nader Akoury, Shufan Wang, Josh Whiting, Stephen Hood, Nanyun Peng, and Mohit
  Iyyer. 2020.
\newblock \href {https://doi.org/10.18653/v1/2020.emnlp-main.525} {{STORIUM}:
  {A} {D}ataset and {E}valuation {P}latform for {M}achine-in-the-{L}oop {S}tory
  {G}eneration}.
\newblock In \emph{Proceedings of the 2020 Conference on Empirical Methods in
  Natural Language Processing (EMNLP)}, pages 6470--6484, Online. Association
  for Computational Linguistics.

\bibitem[{Bamman et~al.(2013)Bamman, O’Connor, and
  Smith}]{bamman2013learning}
David Bamman, Brendan O’Connor, and Noah~A Smith. 2013.
\newblock Learning latent personas of film characters.
\newblock In \emph{Proceedings of the 51st Annual Meeting of the Association
  for Computational Linguistics (Volume 1: Long Papers)}, pages 352--361.

\bibitem[{Bhagavatula et~al.(2019)Bhagavatula, Le~Bras, Malaviya, Sakaguchi,
  Holtzman, Rashkin, Downey, Yih, and Choi}]{bhagavatula2019abductive}
Chandra Bhagavatula, Ronan Le~Bras, Chaitanya Malaviya, Keisuke Sakaguchi, Ari
  Holtzman, Hannah Rashkin, Doug Downey, Wen-tau Yih, and Yejin Choi. 2019.
\newblock Abductive commonsense reasoning.
\newblock In \emph{International Conference on Learning Representations}.

\bibitem[{Brahman and Chaturvedi(2020)}]{brahman2020modeling}
Faeze Brahman and Snigdha Chaturvedi. 2020.
\newblock \href {https://doi.org/10.18653/v1/2020.emnlp-main.426} {Modeling
  protagonist emotions for emotion-aware storytelling}.
\newblock In \emph{Proceedings of the 2020 Conference on Empirical Methods in
  Natural Language Processing (EMNLP)}, pages 5277--5294, Online. Association
  for Computational Linguistics.

\bibitem[{Brown et~al.(2020)Brown, Mann, Ryder, Subbiah, Kaplan, Dhariwal,
  Neelakantan, Shyam, Sastry, Askell, Agarwal, Herbert-Voss, Krueger, Henighan,
  Child, Ramesh, Ziegler, Wu, Winter, Hesse, Chen, Sigler, Litwin, Gray, Chess,
  Clark, Berner, McCandlish, Radford, Sutskever, and
  Amodei}]{brown2020language}
Tom~B. Brown, Benjamin Mann, Nick Ryder, Melanie Subbiah, Jared Kaplan,
  Prafulla Dhariwal, Arvind Neelakantan, Pranav Shyam, Girish Sastry, Amanda
  Askell, Sandhini Agarwal, Ariel Herbert-Voss, Gretchen Krueger, Tom Henighan,
  Rewon Child, Aditya Ramesh, Daniel~M. Ziegler, Jeffrey Wu, Clemens Winter,
  Christopher Hesse, Mark Chen, Eric Sigler, Mateusz Litwin, Scott Gray,
  Benjamin Chess, Jack Clark, Christopher Berner, Sam McCandlish, Alec Radford,
  Ilya Sutskever, and Dario Amodei. 2020.
\newblock \href {http://arxiv.org/abs/2005.14165} {Language models are few-shot
  learners}.

\bibitem[{Chambers and Jurafsky(2008)}]{chambers2008unsupervised}
Nathanael Chambers and Dan Jurafsky. 2008.
\newblock Unsupervised learning of narrative event chains.
\newblock In \emph{Proceedings of ACL-08: HLT}, pages 789--797.

\bibitem[{Chaturvedi et~al.(2017)Chaturvedi, Iyyer, and
  Daume~III}]{chaturvedi2017unsupervised}
Snigdha Chaturvedi, Mohit Iyyer, and Hal Daume~III. 2017.
\newblock Unsupervised learning of evolving relationships between literary
  characters.
\newblock In \emph{Proceedings of the AAAI Conference on Artificial
  Intelligence}, volume~31.

\bibitem[{Chaturvedi et~al.(2016)Chaturvedi, Srivastava, Daume~III, and
  Dyer}]{chaturvedi2016modeling}
Snigdha Chaturvedi, Shashank Srivastava, Hal Daume~III, and Chris Dyer. 2016.
\newblock Modeling evolving relationships between characters in literary
  novels.
\newblock In \emph{Proceedings of the AAAI Conference on Artificial
  Intelligence}, volume~30.

\bibitem[{Chen et~al.(2019)Chen, Chu, and Gimpel}]{chen2019evaluation}
Mingda Chen, Zewei Chu, and Kevin Gimpel. 2019.
\newblock Evaluation benchmarks and learning criteria for discourse-aware
  sentence representations.
\newblock In \emph{Proceedings of the 2019 Conference on Empirical Methods in
  Natural Language Processing and the 9th International Joint Conference on
  Natural Language Processing (EMNLP-IJCNLP)}, pages 649--662.

\bibitem[{Conneau and Kiela(2018)}]{conneau2018senteval}
Alexis Conneau and Douwe Kiela. 2018.
\newblock Senteval: An evaluation toolkit for universal sentence
  representations.
\newblock In \emph{Proceedings of the Eleventh International Conference on
  Language Resources and Evaluation (LREC 2018)}.

\bibitem[{Cui et~al.(2020)Cui, Che, Liu, Qin, Wang, and Hu}]{cui2020revisiting}
Yiming Cui, Wanxiang Che, Ting Liu, Bing Qin, Shijin Wang, and Guoping Hu.
  2020.
\newblock \href {https://www.aclweb.org/anthology/2020.findings-emnlp.58}
  {Revisiting pre-trained models for {C}hinese natural language processing}.
\newblock In \emph{Proceedings of the 2020 Conference on Empirical Methods in
  Natural Language Processing: Findings}, pages 657--668, Online. Association
  for Computational Linguistics.

\bibitem[{Dai et~al.(2019)Dai, Yang, Yang, Carbonell, Le, and
  Salakhutdinov}]{dai2019transformer}
Zihang Dai, Zhilin Yang, Yiming Yang, Jaime~G Carbonell, Quoc Le, and Ruslan
  Salakhutdinov. 2019.
\newblock Transformer-xl: Attentive language models beyond a fixed-length
  context.
\newblock In \emph{Proceedings of the 57th Annual Meeting of the Association
  for Computational Linguistics}, pages 2978--2988.

\bibitem[{Devlin et~al.(2019)Devlin, Chang, Lee, and
  Toutanova}]{devlin2018bert}
Jacob Devlin, Ming-Wei Chang, Kenton Lee, and Kristina Toutanova. 2019.
\newblock Bert: Pre-training of deep bidirectional transformers for language
  understanding.
\newblock In \emph{Proceedings of the 2019 Conference of the North American
  Chapter of the Association for Computational Linguistics: Human Language
  Technologies, Volume 1 (Long and Short Papers)}, pages 4171--4186.

\bibitem[{Fan et~al.(2018)Fan, Lewis, and Dauphin}]{fan2018hierarchical}
Angela Fan, Mike Lewis, and Yann Dauphin. 2018.
\newblock Hierarchical neural story generation.
\newblock In \emph{Proceedings of the 56th Annual Meeting of the Association
  for Computational Linguistics (Volume 1: Long Papers)}, pages 889--898.

\bibitem[{Finlayson(2012)}]{finlayson2012learning}
Mark Mark~Alan Finlayson. 2012.
\newblock \emph{Learning narrative structure from annotated folktales}.
\newblock Ph.D. thesis, Massachusetts Institute of Technology.

\bibitem[{Fleiss and Joseph(1971)}]{Fleiss1971Measuring}
Fleiss and L.~Joseph. 1971.
\newblock Measuring nominal scale agreement among many raters.
\newblock \emph{Psychological Bulletin}, 76(5):378--382.

\bibitem[{Gehring et~al.(2017)Gehring, Auli, Grangier, Yarats, and
  Dauphin}]{gehring2017convolutional}
Jonas Gehring, Michael Auli, David Grangier, Denis Yarats, and Yann~N Dauphin.
  2017.
\newblock Convolutional sequence to sequence learning.
\newblock In \emph{International Conference on Machine Learning}, pages
  1243--1252. PMLR.

\bibitem[{Gehrmann et~al.(2021)Gehrmann, Adewumi, Aggarwal, Ammanamanchi,
  Anuoluwapo, Bosselut, Chandu, Clinciu, Das, Dhole et~al.}]{gehrmann2021gem}
Sebastian Gehrmann, Tosin Adewumi, Karmanya Aggarwal, Pawan~Sasanka
  Ammanamanchi, Aremu Anuoluwapo, Antoine Bosselut, Khyathi~Raghavi Chandu,
  Miruna Clinciu, Dipanjan Das, Kaustubh~D Dhole, et~al. 2021.
\newblock The gem benchmark: Natural language generation, its evaluation and
  metrics.
\newblock \emph{arXiv preprint arXiv:2102.01672}.

\bibitem[{Goodfellow et~al.(2014)Goodfellow, Pouget-Abadie, Mirza, Xu,
  Warde-Farley, Ozair, Courville, and Bengio}]{goodfellow2014generative}
Ian Goodfellow, Jean Pouget-Abadie, Mehdi Mirza, Bing Xu, David Warde-Farley,
  Sherjil Ozair, Aaron Courville, and Yoshua Bengio. 2014.
\newblock Generative adversarial nets.
\newblock In \emph{Advances in neural information processing systems}, pages
  2672--2680.

\bibitem[{Guan et~al.(2020)Guan, Huang, Zhao, Zhu, and
  Huang}]{guan2020knowledge}
Jian Guan, Fei Huang, Zhihao Zhao, Xiaoyan Zhu, and Minlie Huang. 2020.
\newblock A knowledge-enhanced pretraining model for commonsense story
  generation.
\newblock \emph{Transactions of the Association for Computational Linguistics},
  8:93--108.

\bibitem[{Guan and Huang(2020)}]{guan2020union}
Jian Guan and Minlie Huang. 2020.
\newblock \href {https://doi.org/10.18653/v1/2020.emnlp-main.736} {{UNION:} an
  unreferenced metric for evaluating open-ended story generation}.
\newblock In \emph{Proceedings of the 2020 Conference on Empirical Methods in
  Natural Language Processing, {EMNLP} 2020, Online, November 16-20, 2020},
  pages 9157--9166. Association for Computational Linguistics.

\bibitem[{Guan et~al.(2019)Guan, Wang, and Huang}]{guan2019story}
Jian Guan, Yansen Wang, and Minlie Huang. 2019.
\newblock Story ending generation with incremental encoding and commonsense
  knowledge.
\newblock In \emph{Proceedings of the AAAI Conference on Artificial
  Intelligence}, volume~33, pages 6473--6480.

\bibitem[{Guan et~al.(2021)Guan, Zhang, Feng, Liu, Ding, Mao, Fan, and
  Huang}]{guan2021openmeva}
Jian Guan, Zhexin Zhang, Zhuoer Feng, Zitao Liu, Wenbiao Ding, Xiaoxi Mao,
  Changjie Fan, and Minlie Huang. 2021.
\newblock \href {https://doi.org/10.18653/v1/2021.acl-long.500} {{O}pen{MEVA}:
  A benchmark for evaluating open-ended story generation metrics}.
\newblock In \emph{Proceedings of the 59th Annual Meeting of the Association
  for Computational Linguistics and the 11th International Joint Conference on
  Natural Language Processing (Volume 1: Long Papers)}, pages 6394--6407,
  Online. Association for Computational Linguistics.

\bibitem[{Kong et~al.(2021)Kong, Huang, Tung, Guan, and
  Huang}]{kong-etal-2021-stylized}
Xiangzhe Kong, Jialiang Huang, Ziquan Tung, Jian Guan, and Minlie Huang. 2021.
\newblock \href {https://doi.org/10.18653/v1/2021.findings-acl.215} {Stylized
  story generation with style-guided planning}.
\newblock In \emph{Findings of the Association for Computational Linguistics:
  ACL-IJCNLP 2021}, pages 2430--2436, Online. Association for Computational
  Linguistics.

\bibitem[{Kudo and Richardson(2018)}]{kudo2018sentencepiece}
Taku Kudo and John Richardson. 2018.
\newblock Sentencepiece: A simple and language independent subword tokenizer
  and detokenizer for neural text processing.
\newblock In \emph{Proceedings of the 2018 Conference on Empirical Methods in
  Natural Language Processing: System Demonstrations}, pages 66--71.

\bibitem[{Lewis et~al.(2020)Lewis, Liu, Goyal, Ghazvininejad, Mohamed, Levy,
  Stoyanov, and Zettlemoyer}]{bart}
Mike Lewis, Yinhan Liu, Naman Goyal, Marjan Ghazvininejad, Abdelrahman Mohamed,
  Omer Levy, Veselin Stoyanov, and Luke Zettlemoyer. 2020.
\newblock \href {https://www.aclweb.org/anthology/2020.acl-main.703/} {{BART:}
  denoising sequence-to-sequence pre-training for natural language generation,
  translation, and comprehension}.
\newblock In \emph{Proceedings of the 58th Annual Meeting of the Association
  for Computational Linguistics, {ACL} 2020, Online, July 5-10, 2020}, pages
  7871--7880. Association for Computational Linguistics.

\bibitem[{Li et~al.(2013)Li, Lee-Urban, Johnston, and Riedl}]{li2013story}
Boyang Li, Stephen Lee-Urban, George Johnston, and Mark Riedl. 2013.
\newblock Story generation with crowdsourced plot graphs.
\newblock In \emph{Proceedings of the AAAI Conference on Artificial
  Intelligence}, volume~27.

\bibitem[{Li et~al.(2016)Li, Galley, Brockett, Gao, and
  Dolan}]{li2016diversity}
Jiwei Li, Michel Galley, Chris Brockett, Jianfeng Gao, and William~B Dolan.
  2016.
\newblock A diversity-promoting objective function for neural conversation
  models.
\newblock In \emph{Proceedings of the 2016 Conference of the North American
  Chapter of the Association for Computational Linguistics: Human Language
  Technologies}, pages 110--119.

\bibitem[{Lin(2004)}]{lin-2004-rouge}
Chin-Yew Lin. 2004.
\newblock \href {https://www.aclweb.org/anthology/W04-1013} {{ROUGE}: A package
  for automatic evaluation of summaries}.
\newblock In \emph{Text Summarization Branches Out}, pages 74--81, Barcelona,
  Spain. Association for Computational Linguistics.

\bibitem[{Liu et~al.(2020)Liu, Yan, Gong, Qi, Zhang, Jiao, Chen, Fu, Shou, Gong
  et~al.}]{liu2020glge}
Dayiheng Liu, Yu~Yan, Yeyun Gong, Weizhen Qi, Hang Zhang, Jian Jiao, Weizhu
  Chen, Jie Fu, Linjun Shou, Ming Gong, et~al. 2020.
\newblock Glge: A new general language generation evaluation benchmark.
\newblock \emph{arXiv preprint arXiv:2011.11928}.

\bibitem[{Louis and Sutton(2018)}]{louis2018deep}
Annie Louis and Charles Sutton. 2018.
\newblock Deep dungeons and dragons: Learning character-action interactions
  from role-playing game transcripts.
\newblock In \emph{Proceedings of the 2018 Conference of the North American
  Chapter of the Association for Computational Linguistics: Human Language
  Technologies, Volume 2 (Short Papers)}, pages 708--713.

\bibitem[{Merity et~al.(2016)Merity, Xiong, Bradbury, and
  Socher}]{merity2016pointer}
Stephen Merity, Caiming Xiong, James Bradbury, and Richard Socher. 2016.
\newblock Pointer sentinel mixture models.
\newblock \emph{arXiv preprint arXiv:1609.07843}.

\bibitem[{Mostafazadeh et~al.(2016)Mostafazadeh, Chambers, He, Parikh, Batra,
  Vanderwende, Kohli, and Allen}]{mostafazadeh2016corpus}
Nasrin Mostafazadeh, Nathanael Chambers, Xiaodong He, Devi Parikh, Dhruv Batra,
  Lucy Vanderwende, Pushmeet Kohli, and James Allen. 2016.
\newblock A corpus and cloze evaluation for deeper understanding of commonsense
  stories.
\newblock In \emph{Proceedings of NAACL-HLT}, pages 839--849.

\bibitem[{Papineni et~al.(2002)Papineni, Roukos, Ward, and
  Zhu}]{papineni2002bleu}
Kishore Papineni, Salim Roukos, Todd Ward, and Wei-Jing Zhu. 2002.
\newblock Bleu: a method for automatic evaluation of machine translation.
\newblock In \emph{Proceedings of the 40th annual meeting of the Association
  for Computational Linguistics}, pages 311--318.

\bibitem[{Paul and Frank(2021)}]{paul-frank-2021-coins}
Debjit Paul and Anette Frank. 2021.
\newblock \href {https://doi.org/10.18653/v1/2021.acl-long.395} {{COINS}:
  Dynamically generating {CO}ntextualized inference rules for narrative story
  completion}.
\newblock In \emph{Proceedings of the 59th Annual Meeting of the Association
  for Computational Linguistics and the 11th International Joint Conference on
  Natural Language Processing (Volume 1: Long Papers)}, pages 5086--5099,
  Online. Association for Computational Linguistics.

\bibitem[{Radford et~al.(2018)Radford, Narasimhan, Salimans, and
  Sutskever}]{radford2018improving}
Alec Radford, Karthik Narasimhan, Tim Salimans, and Ilya Sutskever. 2018.
\newblock Improving language understanding with unsupervised learning.

\bibitem[{Radford et~al.(2019)Radford, Wu, Child, Luan, Amodei, and
  Sutskever}]{radford2019language}
Alec Radford, Jeffrey Wu, Rewon Child, David Luan, Dario Amodei, and Ilya
  Sutskever. 2019.
\newblock Language models are unsupervised multitask learners.
\newblock \emph{OpenAI blog}, 1(8):9.

\bibitem[{Rae et~al.(2020)Rae, Potapenko, Jayakumar, Hillier, and
  Lillicrap}]{Rae2020Compressive}
Jack~W. Rae, Anna Potapenko, Siddhant~M. Jayakumar, Chloe Hillier, and
  Timothy~P. Lillicrap. 2020.
\newblock \href {https://openreview.net/forum?id=SylKikSYDH} {Compressive
  transformers for long-range sequence modelling}.
\newblock In \emph{International Conference on Learning Representations}.

\bibitem[{Raffel et~al.(2020)Raffel, Shazeer, Roberts, Lee, Narang, Matena,
  Zhou, Li, and Liu}]{raffel2020exploring}
Colin Raffel, Noam Shazeer, Adam Roberts, Katherine Lee, Sharan Narang, Michael
  Matena, Yanqi Zhou, Wei Li, and Peter~J Liu. 2020.
\newblock Exploring the limits of transfer learning with a unified text-to-text
  transformer.
\newblock \emph{Journal of Machine Learning Research}, 21:1--67.

\bibitem[{Rashkin et~al.(2020)Rashkin, Celikyilmaz, Choi, and
  Gao}]{rashkin2020plotmachines}
Hannah Rashkin, Asli Celikyilmaz, Yejin Choi, and Jianfeng Gao. 2020.
\newblock Plotmachines: Outline-conditioned generation with dynamic plot state
  tracking.
\newblock In \emph{Proceedings of the 2020 Conference on Empirical Methods in
  Natural Language Processing (EMNLP)}, pages 4274--4295.

\bibitem[{Ribeiro et~al.(2020)Ribeiro, Wu, Guestrin, and
  Singh}]{ribeiro-etal-2020-beyond}
Marco~Tulio Ribeiro, Tongshuang Wu, Carlos Guestrin, and Sameer Singh. 2020.
\newblock \href {https://doi.org/10.18653/v1/2020.acl-main.442} {Beyond
  accuracy: Behavioral testing of {NLP} models with {C}heck{L}ist}.
\newblock In \emph{Proceedings of the 58th Annual Meeting of the Association
  for Computational Linguistics}, pages 4902--4912, Online. Association for
  Computational Linguistics.

\bibitem[{Rockt{\"{a}}schel et~al.(2016)Rockt{\"{a}}schel, Grefenstette,
  Hermann, Kocisk{\'{y}}, and Blunsom}]{DBLP:journals/corr/RocktaschelGHKB15}
Tim Rockt{\"{a}}schel, Edward Grefenstette, Karl~Moritz Hermann, Tom{\'{a}}s
  Kocisk{\'{y}}, and Phil Blunsom. 2016.
\newblock \href {http://arxiv.org/abs/1509.06664} {Reasoning about entailment
  with neural attention}.
\newblock In \emph{4th International Conference on Learning Representations,
  {ICLR} 2016, San Juan, Puerto Rico, May 2-4, 2016, Conference Track
  Proceedings}.

\bibitem[{Rose et~al.(2010)Rose, Engel, Cramer, and Cowley}]{rose2010automatic}
Stuart Rose, Dave Engel, Nick Cramer, and Wendy Cowley. 2010.
\newblock Automatic keyword extraction from individual documents.
\newblock \emph{Text mining: applications and theory}, 1:1--20.

\bibitem[{Sarlin et~al.(2020)Sarlin, DeTone, Malisiewicz, and
  Rabinovich}]{sarlin2020superglue}
Paul-Edouard Sarlin, Daniel DeTone, Tomasz Malisiewicz, and Andrew Rabinovich.
  2020.
\newblock Superglue: Learning feature matching with graph neural networks.
\newblock In \emph{Proceedings of the IEEE/CVF conference on computer vision
  and pattern recognition}, pages 4938--4947.

\bibitem[{Schwartz et~al.(2017)Schwartz, Sap, Konstas, Zilles, Choi, and
  Smith}]{schwartz2017effect}
Roy Schwartz, Maarten Sap, Ioannis Konstas, Leila Zilles, Yejin Choi, and
  Noah~A Smith. 2017.
\newblock The effect of different writing tasks on linguistic style: A case
  study of the roc story cloze task.
\newblock In \emph{Proceedings of the 21st Conference on Computational Natural
  Language Learning (CoNLL 2017)}, pages 15--25.

\bibitem[{Sharma et~al.(2018)Sharma, Allen, Bakhshandeh, and
  Mostafazadeh}]{sharma2018tackling}
Rishi Sharma, James Allen, Omid Bakhshandeh, and Nasrin Mostafazadeh. 2018.
\newblock Tackling the story ending biases in the story cloze test.
\newblock In \emph{Proceedings of the 56th Annual Meeting of the Association
  for Computational Linguistics (Volume 2: Short Papers)}, pages 752--757.

\bibitem[{Tay et~al.(2020)Tay, Dehghani, Abnar, Shen, Bahri, Pham, Rao, Yang,
  Ruder, and Metzler}]{tay2020long}
Yi~Tay, Mostafa Dehghani, Samira Abnar, Yikang Shen, Dara Bahri, Philip Pham,
  Jinfeng Rao, Liu Yang, Sebastian Ruder, and Donald Metzler. 2020.
\newblock Long range arena: A benchmark for efficient transformers.
\newblock In \emph{International Conference on Learning Representations}.

\bibitem[{Vaswani et~al.(2017)Vaswani, Shazeer, Parmar, Uszkoreit, Jones,
  Gomez, Kaiser, and Polosukhin}]{vaswani2017attention}
Ashish Vaswani, Noam Shazeer, Niki Parmar, Jakob Uszkoreit, Llion Jones,
  Aidan~N Gomez, {\L}ukasz Kaiser, and Illia Polosukhin. 2017.
\newblock Attention is all you need.
\newblock In \emph{Advances in neural information processing systems}, pages
  5998--6008.

\bibitem[{Wang et~al.(2019)Wang, Singh, Michael, Hill, Levy, and
  Bowman}]{wang2018glue}
Alex Wang, Amanpreet Singh, Julian Michael, Felix Hill, Omer Levy, and
  Samuel~R. Bowman. 2019.
\newblock \href {https://openreview.net/forum?id=rJ4km2R5t7} {{GLUE}: A
  multi-task benchmark and analysis platform for natural language
  understanding}.
\newblock In \emph{International Conference on Learning Representations}.

\bibitem[{Wang and Wan(2019)}]{DBLP:conf/ijcai/Wang019b}
Tianming Wang and Xiaojun Wan. 2019.
\newblock \href {https://doi.org/10.24963/ijcai.2019/727} {{T-CVAE:}
  transformer-based conditioned variational autoencoder for story completion}.
\newblock In \emph{Proceedings of the Twenty-Eighth International Joint
  Conference on Artificial Intelligence, {IJCAI} 2019, Macao, China, August
  10-16, 2019}, pages 5233--5239. ijcai.org.

\bibitem[{Xu et~al.(2020{\natexlab{a}})Xu, Hu, Zhang, Li, Cao, Li, Xu, Sun, Yu,
  Yu et~al.}]{xu2020clue}
Liang Xu, Hai Hu, Xuanwei Zhang, Lu~Li, Chenjie Cao, Yudong Li, Yechen Xu, Kai
  Sun, Dian Yu, Cong Yu, et~al. 2020{\natexlab{a}}.
\newblock Clue: A chinese language understanding evaluation benchmark.
\newblock In \emph{Proceedings of the 28th International Conference on
  Computational Linguistics}, pages 4762--4772.

\bibitem[{Xu et~al.(2020{\natexlab{b}})Xu, Patwary, Shoeybi, Puri, Fung,
  Anandkumar, and Catanzaro}]{DBLP:conf/emnlp/XuPSPFAC20}
Peng Xu, Mostofa Patwary, Mohammad Shoeybi, Raul Puri, Pascale Fung, Anima
  Anandkumar, and Bryan Catanzaro. 2020{\natexlab{b}}.
\newblock \href {https://doi.org/10.18653/v1/2020.emnlp-main.226}
  {{MEGATRON-CNTRL:} controllable story generation with external knowledge
  using large-scale language models}.
\newblock In \emph{Proceedings of the 2020 Conference on Empirical Methods in
  Natural Language Processing, {EMNLP} 2020, Online, November 16-20, 2020},
  pages 2831--2845. Association for Computational Linguistics.

\bibitem[{Xue et~al.(2021)Xue, Constant, Roberts, Kale, Al-Rfou, Siddhant,
  Barua, and Raffel}]{xue2021mt5}
Linting Xue, Noah Constant, Adam Roberts, Mihir Kale, Rami Al-Rfou, Aditya
  Siddhant, Aditya Barua, and Colin Raffel. 2021.
\newblock mt5: A massively multilingual pre-trained text-to-text transformer.
\newblock In \emph{Proceedings of the 2021 Conference of the North American
  Chapter of the Association for Computational Linguistics: Human Language
  Technologies}, pages 483--498.

\bibitem[{Yao et~al.(2019)Yao, Peng, Weischedel, Knight, Zhao, and
  Yan}]{yao2019plan}
Lili Yao, Nanyun Peng, Ralph Weischedel, Kevin Knight, Dongyan Zhao, and Rui
  Yan. 2019.
\newblock Plan-and-write: Towards better automatic storytelling.
\newblock In \emph{Proceedings of the AAAI Conference on Artificial
  Intelligence}, volume~33, pages 7378--7385.

\bibitem[{Zhang et~al.(2020)Zhang, Han, Zhou, Ke, Gu, Ye, Qin, Su, Ji, Guan
  et~al.}]{zhang2020cpm}
Zhengyan Zhang, Xu~Han, Hao Zhou, Pei Ke, Yuxian Gu, Deming Ye, Yujia Qin,
  Yusheng Su, Haozhe Ji, Jian Guan, et~al. 2020.
\newblock Cpm: A large-scale generative chinese pre-trained language model.
\newblock \emph{arXiv preprint arXiv:2012.00413}.

\bibitem[{Zhao et~al.(2017)Zhao, Zhao, and Eskenazi}]{zhao2017learning}
Tiancheng Zhao, Ran Zhao, and Maxine Eskenazi. 2017.
\newblock Learning discourse-level diversity for neural dialog models using
  conditional variational autoencoders.
\newblock In \emph{Proceedings of the 55th Annual Meeting of the Association
  for Computational Linguistics (Volume 1: Long Papers)}, pages 654--664.

\bibitem[{Zhao et~al.(2019)Zhao, Chen, Zhang, Zhao, Liu, Lu, Chen, Deng, Ju,
  and Du}]{zhao2019uer}
Zhe Zhao, Hui Chen, Jinbin Zhang, Xin Zhao, Tao Liu, Wei Lu, Xi~Chen, Haotang
  Deng, Qi~Ju, and Xiaoyong Du. 2019.
\newblock Uer: An open-source toolkit for pre-training models.
\newblock \emph{EMNLP-IJCNLP 2019}, page 241.

\end{thebibliography}
\bibliographystyle{acl_natbib}

\end{document}